\documentclass[journal]{IEEEtran}
\ifCLASSINFOpdf
\else
\fi
\usepackage{algorithmic}
\usepackage{url}

\usepackage{mathrsfs}
\usepackage{multirow}
\usepackage{cite}
\usepackage{verbatim}
\usepackage{amsmath,amssymb,amsfonts}
\usepackage{lineno,hyperref}
\usepackage{stfloats}
\usepackage{float}
\usepackage{graphicx}
\usepackage{subfigure}
\usepackage{epstopdf}
\usepackage{textcomp}
\usepackage{bm}
\usepackage{color} 
\usepackage[english]{babel}
\usepackage{blindtext}
\usepackage[linesnumbered,ruled,lined,boxed]{algorithm2e}

\begin{document}
%
\title{An Order-Invariant and Interpretable Hierarchical Dilated Convolution Neural Network for Chemical Fault Detection and Diagnosis}
%
%
%

\author{Mengxuan Li,
        Peng Peng, \IEEEmembership{Member, IEEE,}
        Min Wang, \IEEEmembership{Member, IEEE,}
        Hongwei Wang

\thanks{*This work was supported by the National Key R\&D Program of China under Grant 2020YFB1707803.(Corresponding author: Peng Peng and Hongwei Wang.)}
\thanks{
Mengxuan Li is with the College of Computer Science and Technology in Zhejiang University, Hangzhou, 310013, China. (E-mail: mengxuanli@intl.zju.edu.cn).

Peng Peng and Hongwei Wang are with Zhejiang University and the University of Illinois Urbana–Champaign Institute, Haining, 314400, China. (E-mail: pengpeng@intl.zju.edu.cn, hongweiwang@intl.zju.edu.cn).

Min Wang is with the School of Automation Engineering, University of Electronic Science and Technology of China, Chengdu 611731, China, (e-mail: mwang@uestc.edu.cn).
}
\thanks{This work has been submitted to the IEEE Transactions on Automation Science and Engineering for possible publication. Copyright may be transferred without notice, after which this version may no longer be accessible.}

}

\maketitle

\begin{abstract}
Fault detection and diagnosis is significant for reducing maintenance costs and improving health and safety in chemical processes. Convolution neural network (CNN) is a popular deep learning algorithm with many successful applications in chemical fault detection and diagnosis tasks. However, convolution layers in CNN are very sensitive to the order of features, which can lead to instability in the processing of tabular data. Optimal order of features result in better performance of CNN models but it is expensive to seek such optimal order. In addition, because of the encapsulation mechanism of feature extraction, most CNN models are opaque and have poor interpretability, thus failing to identify root-cause features without human supervision. These difficulties inevitably limit the performance and credibility of CNN methods. In this paper, we propose an order-invariant and interpretable hierarchical dilated convolution neural network (HDLCNN), which is composed by feature clustering, dilated convolution and the shapley additive explanations (SHAP) method. The novelty of HDLCNN lies in its capability of processing tabular data with features of arbitrary order without seeking the optimal order, due to the ability to agglomerate correlated features of feature clustering and the large receptive field of dilated convolution. Then, the proposed method provides interpretability by including the SHAP values to quantify feature contribution. Therefore, the root-cause features can be identified as the features with the highest contribution. Computational experiments are conducted on the Tennessee Eastman chemical process benchmark dataset. Compared with the other methods, the proposed HDLCNN-SHAP method achieves better performance on processing tabular data with features of arbitrary order, detecting faults, and identifying the root-cause features.

\end{abstract}

\def\abstractname{Note to Practitioners}
\begin{abstract}
This paper was motivated by the problem of fault detection and diagnosis to process multiple variables and identify the root-cause features in real chemical processes. In this case, the order of features affects the fault detection performance and the precise root-cause feature is required to be identify to avoid the same faults. This paper presents a novel order-invariant and interpretable framework for fault detection and root cause analysis in real chemical processes to process data with features of arbitrary order, thus reducing the burden on users. It utilizes the collected historical data for training and requires no human supervision. The newly collected data is automatically classified into a normal or fault type. Once a fault happens, the corresponding root-cause feature is identified without any prior knowledge. In our future work, we plan to focus on the faults with multiple root-cause features. In addition, the incomplete dataset and simultaneous-fault diagnosis are worth of investigation.
\end{abstract}


%
\IEEEpeerreviewmaketitle

\section{Introduction}
%
%
%
%

\IEEEPARstart{w}ITH the advent of Industry 4.0, chemical processes become more intelligent and automatic. This trend has raised the urgent need of detecting anomalies and diagnosing faults efficiently and correctly. Chemical faults result in chemical contamination, potential explosion, and other seriously chemical hazards, and thus intelligent fault diagnosis methods are required to find the underlying causes of the faults. Current fault detection and diagnosis methods can be classified into two categories: model-based and data-based. Model-based methods are less accurate with higher level of complexity as they depend on the modeling of complex physical and chemical processes. Therefore, data-based methods have become increasingly popular recently. Among these methods, deep learning methods have been used widely and achieved electrifying performance in fault detection and diagnosis problems \cite{xie2021intelligent}.

In particular, convolution neural network (CNN) is one of the most representative deep learning architectures based on convolution calculations. The main advantage of CNN is that it automatically detects the important features without any human supervision. And it can easily process high-dimensional data by sharing convolution kernels. Currently, some work has been done to show the potential of utilizing CNN to detect fault in chemical processes. For example, Wang \emph{et al.} proposed a feature fusion fault diagnosis method using a normalized CNN for complex chemical processes \cite{wang2021fault}. Huang \emph{et al.} introduced a novel fault diagnosis method that consists of sliding window processing and a CNN model \cite{huang2022novel}. However, applying the current CNN models in real chemical processes is still challenging. On the one hand, CNN models rely on convolution operation to extract information within each size-fixed convolution kernel, leading to result that only the information among adjacent features can be extracted. Different from images, the data of chemical processes involves multiple variables through the time domain, which is considered as tabular data. Thus the order of these variables determine the extracted information through kernels, resulting in instability of processing tabular data by convolution layers. On the other hand, the existing CNN methods only provide classification results while the analysis of the root-cause features is lacking. Because of the encapsulation mechanism of feature extraction, most CNN models are opaque without any knowledge of the internal working principles, thus users have no information on feature contribution to the prediction. Without analyzing the underlying root-cause features, the same faults will repeat and result in serious consequences.

In this paper, we propose an order-invariant and interpretable hierarchical dilated convolution neural network (HDLCNN) composed by feature clustering, dilated convolution and shapley additive explanations (SHAP) method to process tabular data with features of arbitrary order and obtain credible root-cause features. Dilated convolution, a variant of CNN that expands the kernel by inserting holes between the kernel elements, is utilized in our method \cite{yu2017dilated}. It is adopted to increase the receptive field size without increasing the number of model parameters. Since receptive field is the corresponding region in the input that determines a unit in a certain layer in the network, dilated convolution with larger receptive field has the capability of extracting global features thus weakening the impact of the order of features. To further eliminate the effects of the order, we utilize feature clustering method to agglomerate highly correlated features before convolution layers. Hierarchical clustering method is applied since it builds a hierarchy of clusters and is not affected by the input order of data. In addition, a major difficulty to apply CNN methods in real chemical processes is to get credible fault detection results and obtain exact root-cause features. This means that interpretability, the degree to which a human can understand the model's result, is vital for human to trust the decisions made by complex models. To solve this problem, we apply the SHAP method to interpret the complex black box model, which is a method to explain individual predictions based on the game theoretically optimal Shapley values \cite{lundberg2017unified-SHAP}. Compared with other interpretability methods, it has the advantages of solid theoretical foundation in game theory and intuitive visualization based on the origin data. Also, it is model-agnostic while providing both local and global interpretability. Therefore, we utilize the SHAP method to provide interpretability by computing the SHAP values to quantify feature contribution and then obtaining the root-cause features.

The main contributions of this article are as follows:

\begin{itemize}
    \item A dilated convolution based order-invariant classifier, namely HDLCNN, is developed to solve chemical fault detection and diagnosis. Dilated convolution is a variant of CNN with larger receptive field, enabling the proposed method to extract more information within a size-fixed convolution kernel thus achieving better performance on processing tabular data with features of arbitrary order.
    \item Hierarchical clustering algorithm is applied to agglomerate highly correlated features before the convolution layers to further weaken the effect of the order of features. As a data pre-processing method, hierarchical clustering treats each input as a separate cluster and then sequentially merges similar clusters thus being unaffected on the order.
    \item SHAP method is applied to provide credible and visual interpretability of the classification results from HDLCNN. The computed SHAP values quantify feature contribution and are utilized to obtain the root-cause features.
    \item The experimental results on the Tennessee Eastman (TE) chemical process benchmark dataset demonstrate that, in contrast to the existing methods, the proposed HDLCNN-SHAP method achieves better performance in the key operations of processing tabular data with features of arbitrary order, detecting faults, and identifying the root-cause features.
\end{itemize}

The rest of this paper is organized as follows. The related work is introduced and the motivation of this work is described in Section \ref{background}. The proposed order-invariant and interpretable chemical fault detection and diagnosis method is shown in Section \ref{method}. The experiments of our proposed method based on the TE dataset are introduced in Section \ref{experiment}. And Section \ref{conclusion} summarizes this paper.

\section{Background Theory and Motivation}
\label{background}
\subsection{Related Work}
Till now, researchers has shown the effectiveness of applying CNN for chemical fault detection and diagnosis. For example, Chadha \emph{et al.} proposed a 1-D CNN model to extract meaningful features from the time series sequence \cite{chadha2019time}. Wang \emph{et al.} proposed a fault diagnosis method using deep learning multi-model fusion based on CNN and long short-term memory (LSTM) \cite{wang2020intelligent}. Gu \emph{et al.} proposed an incremental CNN model to detect faults in a real chemical industrial process \cite{gu2021imbalance}. He \emph{et al.} proposed a multi-block temporal convolutional network to learn the temporal-correlated features \cite{he2021multiblock-mbtcn}. However, these methods mainly focus on extracting temporal features and ignore the effect of the order of features. Zhong \emph{et al.} discussed the impact of the arrangement order of features on fault diagnosis and used enumeration method to find the optimal order \cite{zhong2019novel-optimal-order}. However, it is time consuming to find the optimal order and the problem will be exacerbated since chemical processes involve multiple variables. Therefore, it is necessary to design an effective network that is less affected by the order of features. In this paper, we propose an order-invariant fault detection method based on hierarchical clustering and dilated convolution. The proposed HDLCNN methods can process chemical tabular data with arbitrary feature order and provide accurate fault classification results.

Another limitation of the current CNN methods is that only the classification results are provided without an explanation of the results. Considered as black box systems, the CNN models produces useful fault classification results without revealing any information about its internal workings. To avoid the black box problem, interpretability methods have aroused interests of researchers, which aim to help humans readily understand the reasoning behind predictions made by the complex models. Interpretability methods can be divided into two categories: ante-hoc interpretability and post-hoc interpretability. The former one refers to design interpretable models to directly visualize and interpret the internal information while the latter one refers to apply interpretation methods after model training. Generally, researchers prefer to use ante-hoc interpretable bayesian network (BN) to recognize the root-cause features in chemical processes. BN is a probabilistic graphical model based on random variables and the corresponding conditional probability. It can be used to identify the propagation probability among measurable variables to determine the root-cause features. Liu \emph{et al.} proposed a strong relevant mechanism bayesian network by combining mechanism correlation analysis and process state transition to identify the unmonitored root-cause features \cite{liu2022fault-bn}. Liu \emph{et al.} proposed a multi-state BN to recognize a node into multiple states \cite{liu2022optimized-bn}. However, it is expensive to design a BN model since it requires prior knowledge and expert rules. And no universally acknowledged method is developed for constructing networks from raw data. On the contrary, CNN models provide accurate classification results without human supervision. Therefore, utilizing suitable post-hoc inpretability methods to explain the internal parameters of CNN models is a possible solution to visualize the feature contribution and thus identify the root-cause features. In this paper, we propose an interpretable fault diagnosis method based on the SHAP method. The root-cause features are obtained with SHAP values thus greatly improving the practicability of fault diagnosis methods in real chemical processes.

\begin{figure}[]
\centering
\includegraphics[width = 8cm]{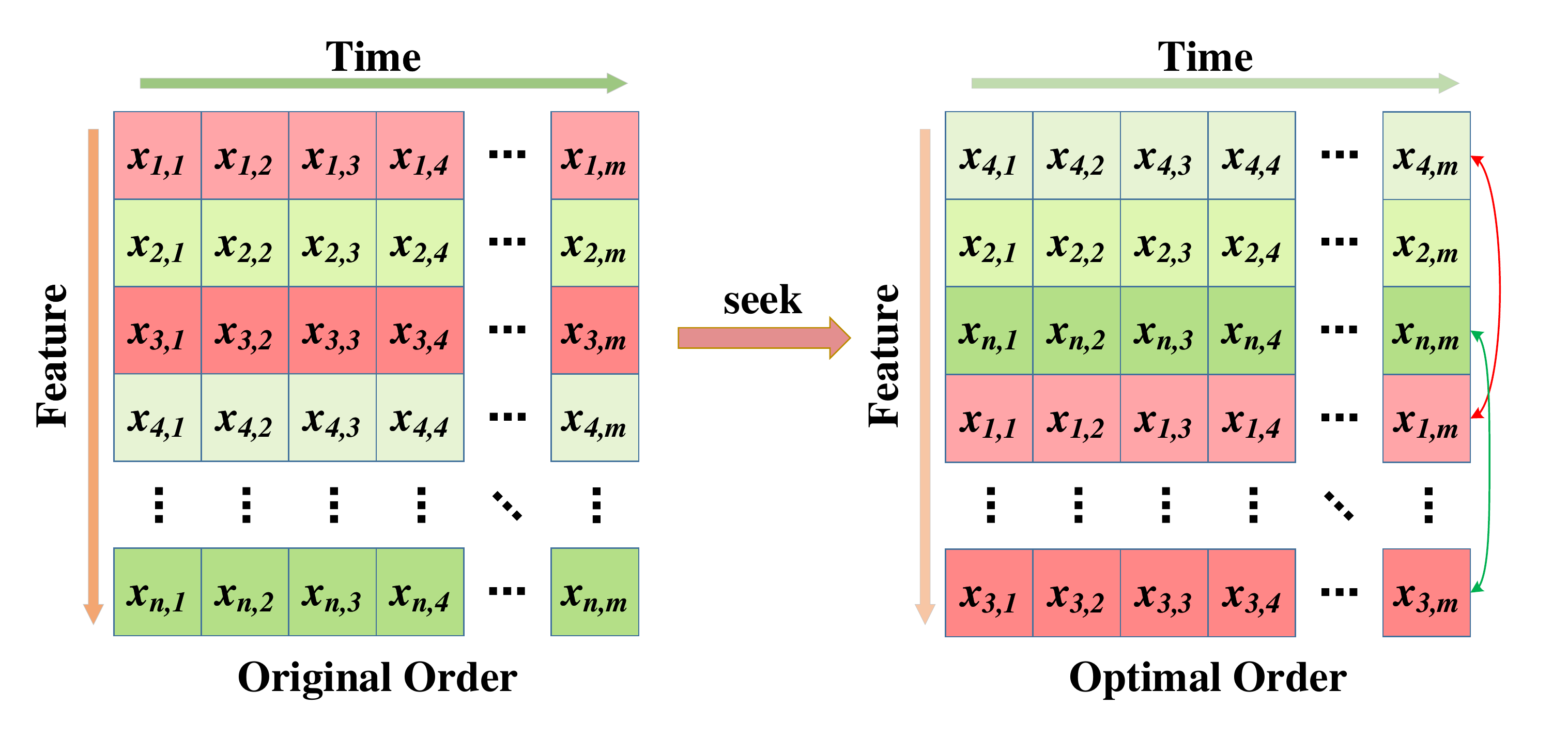}
\caption{The tabular data with $n$ features and $m$ time duration. Highly correlated features are closer in the optimal order. 
}
\label{fig_tabular}
\end{figure}

\subsection{Motivation of This Work}

\begin{figure}[]
\centering
\includegraphics[width = 7cm]{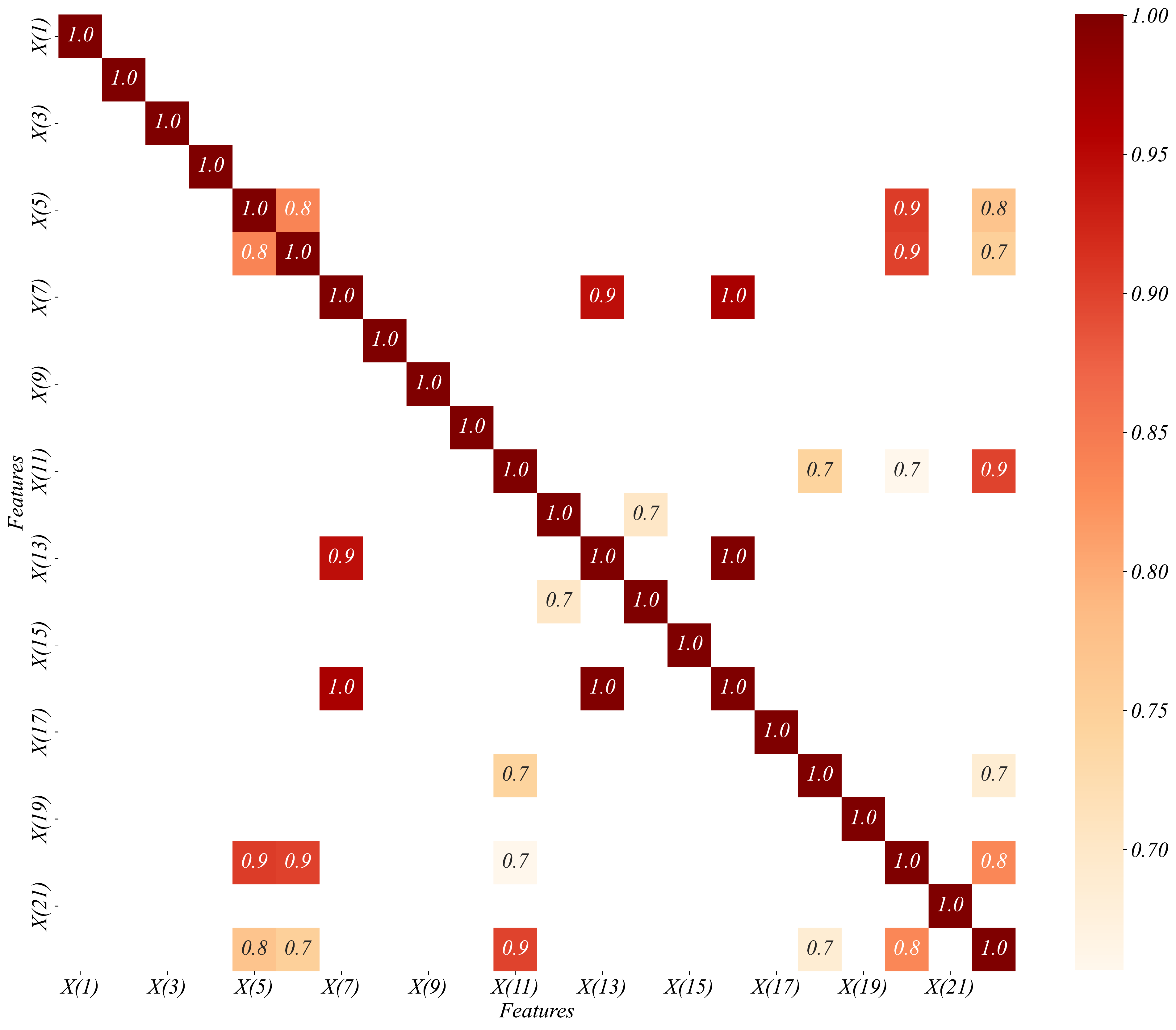}
\caption{The correlation coefficients of the 22 features in the TE dataset.}
\label{fig_corr}
\end{figure}


Fault detection and diagnosis technologies are vital in real chemical processes since they aim to discover the faults in the early stage and thus reduce maintenance costs. For this task, the current CNN methods only focus on extracting temporal features in multi-variables chemical processes. However, chemical processes have tabular data involving both time and feature domains. Similar to the effect of the pixel positions in images, the order of features in tabular data also affect the classification results, since it determines the information extracted within a size-fixed convolution kernel. Although deep CNN models can achieve global receptive field and extract order-invariant information ultimately, they are computational expensive and inefficient for chemical data. Therefore, we aim to design a shallow and order-invariant CNN model. Fig. \ref{fig_tabular} shows an example of tabular data with $n$ features and $m$ time duration. Each row represents a feature and the order of these rows determine the information extracted within a kernel. The left part shows the raw data with original order of features, and the right part shows the optimal order by changing the order of features to make highly correlated features closer. We seek the optimal order since it results in the best performance of CNN model. To further explain this problem, we conduct some experiments on the TE dataset to analyze the correlation of these features. The TE dataset contains 22 continuous process features and the corresponding correlation coefficients are shown in Fig. \ref{fig_corr}. From Fig. \ref{fig_corr}, we can see that there are 14 pairs of features have correlation coefficients greater than 0.7. Therefore, utilizing convolution kernels to extract features of the original order will loss the information related to the correlation of features, since the close-related variables are not considered effectively. Zhong \emph{et al.} proved the impact of the order led on model performance and devoted much efforts to find the optimal order \cite{zhong2019novel-optimal-order}. However, using enumeration method to select the optimal order is inefficient. An alternative solution is to design a model which is less affected by the order of features and can process arbitrary order effectively. In this paper, we propose an order-invariant HDLCNN model based on feature clustering and dilated convolution to extract features with arbitrary order. Dilated convolution is utilized to extract information involving more variables due to its larger receptive field. In addition, as a data pre-processing method, feature clustering is used to further agglomerate correlated features before training the dilated convolution model. More details are described in Section \ref{method}. This design enables the proposed model to effectively process the tabular data with features of arbitrary order thus no longer necessary to seek the optimal order.


On the other hand, chemical processes have a very high risk of serious incident consequences by handling and processing materials under hazardous conditions. Therefore, once a fault is detected, it is necessary to analyze the corresponding root causes to identify the underlying issues and avoid the same faults. Generally, researchers design ante-hoc interpretable BN models to identify root-cause features. But these methods require prior knowledge and expert rules leading to much manual intervention for real applications. On the contrary, post-hoc interpretability methods analyze the complex models after training and require no prior knowledge, thus can be combined with opaque CNN models to leverage their strengths of automatically extracting the important features. In particular, specific post-hoc interpretability methods are proposed to explain the CNN-based models and make them more transparent. Zhou \emph{et al.} utilized global average pooling layers in CNN models to generate class activation maps (CAM), which indicates the discriminative regions used by the CNNs for prediction \cite{zhou2016learning-cam}. Further, Selvaraju \emph{et al.} extends the CAM method to any CNN-based differentiable architecture by using the gradients of targets and flowing into the final convolution layer, namely gradient-weighted CAM (Grad-CAM) \cite{selvaraju2017grad-cam}. However, these CAM-based methods only produce a coarse localization map highlighting the important regions \cite{selvaraju2017grad-cam} while fault diagnosis requires to provide pixel-level explanation and precisely locate the correct root-cause feature. Therefore, these model-specific interpretability methods are not satisfying. In contrast, model-agnostic methods are more flexible and independent of the underlying machine learning model. For example, partial dependence plot (PDP) \cite{friedman2001greedy-pdp} and individual conditional expectation plot (ICEP) \cite{goldstein2015peeking-icep} are designed to display the effect of a feature on the prediction. However, they assume the independence of each feature and fail to process multiple features simultaneously, thus they are inappropriate for chemical processes. Besides, Ribeiro \emph{et al.} proposed a technique that explains individual predictions by training local surrogate models to approximate the predictions of the underlying black box model, namely local interpretable model-agnostic explanations (LIME) \cite{ribeiro2016should-LIME}. But it also ignores the correlation between features and only provide local explanations. In contrast, SHAP provides both local and global explanations by computing the contribution of each feature to the corresponding prediction \cite{lundberg2017unified-SHAP}. Also, it considers the interaction effect after obtaining the individual feature effects. Therefore, we apply SHAP method to improve the interpreability of our CNN-based model. The visualization of feature contribution and analysis of root-cause features based on SHAP values are shown in Section \ref{experiment}.



\subsection{Dilated Convolution}
\begin{figure}[]
\centering
\subfigure[CNN]
{
 \centering
 \includegraphics[width=3.5cm]{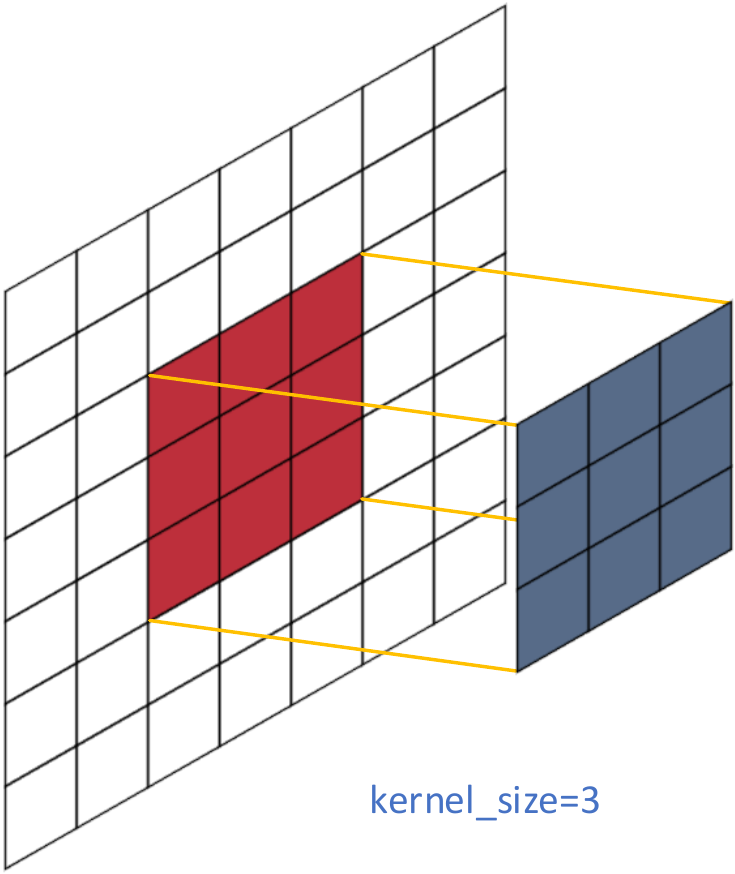}
}
\subfigure[DLCNN]
{
 \centering
 \includegraphics[width=3.5cm]{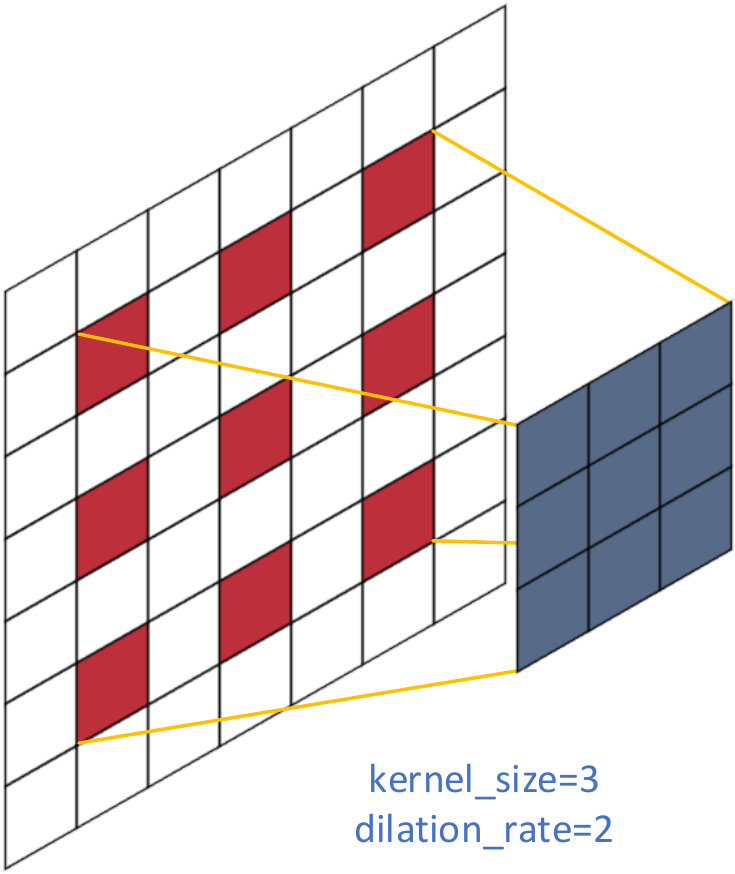}
}
\caption{The comparison of CNN and DLCNN.} 
\label{fig_CNN_dilated}
\end{figure}

CNN is a representative deep learning model which has been widely used in different fields. It takes the raw data, trains the model, then extracts the features automatically for better classification. Although increasing depth of CNN models can achieve larger receptive field size and higher performance, the number of parameters will greatly increase. With the consideration of it, dilated convolution is proposed. The key idea of dilated CNN (DLCNN) is to maintain the high resolution of feature maps and enlarge the receptive field size in CNN \cite{yu2017dilated}. It expands the kernel by inserting holes among original elements thus enlarging the receptive field. Comparing with traditional CNN, it involves a hyper-parameter named dilation rate which indicates a spacing between the non-zero values in a kernel. Fig. \ref{fig_CNN_dilated} shows the comparison of CNN and DLCNN. Theoretically, CNN can be seen as a DLCNN with dilation rate $r = 1$ and the normal convolution calculation is as follows:

\begin{equation}
    Y[i,j] = \sum\nolimits_{m+n=i}\sum\nolimits_{p+q=j}H[m,n] \cdot X[p,q]
\label{CNN_equ}
\end{equation}
where $Y$, $H$ and $X$ are the 2-D output, filter and input respectively. A sample of DLCNN is introduced in Fig. \ref{fig_CNN_dilated} (b). With a dilation rate $r$, ($r - 1$) data points will be skipped in the process of convolution. The dilated convolution calculation is defined as follows:

\begin{equation}
    Y_r[i,j] = \sum\nolimits_{rm+n=i}\sum\nolimits_{rp+q=j}H[m,n] \cdot X[p,q]
\label{DCNN_equ}
\end{equation}
With a dilation rate $r$, this method offers a larger receptive field at the same computational cost. While the number of weights in the kernel is unchanged, they are no longer applied to spatially adjacent samples. Define $c_{l}$ as the receptive field size of the feature map $y_{l}$ at the layer $l$ in a DLCNN. Then, the receptive field size of layer $l$ can be computed as follows:
\begin{equation}
    c_{l} = s_{l+1} \cdot c_{l+1} + [(r_{l+1}(h_{l+1} - 1) + 1) - s_{l+1}]
\label{DCNN_receptice_equ}
\end{equation}
where $s_{l}$ refers to the stride and $h_{l}$ indicates the kernel size. It is obvious that the receptive field increases linearly as the dilation rate increases. This enables models to have a larger receptive field with the same number of parameters and computation costs.

\begin{figure*}[]
\centering
\includegraphics[width = \textwidth]{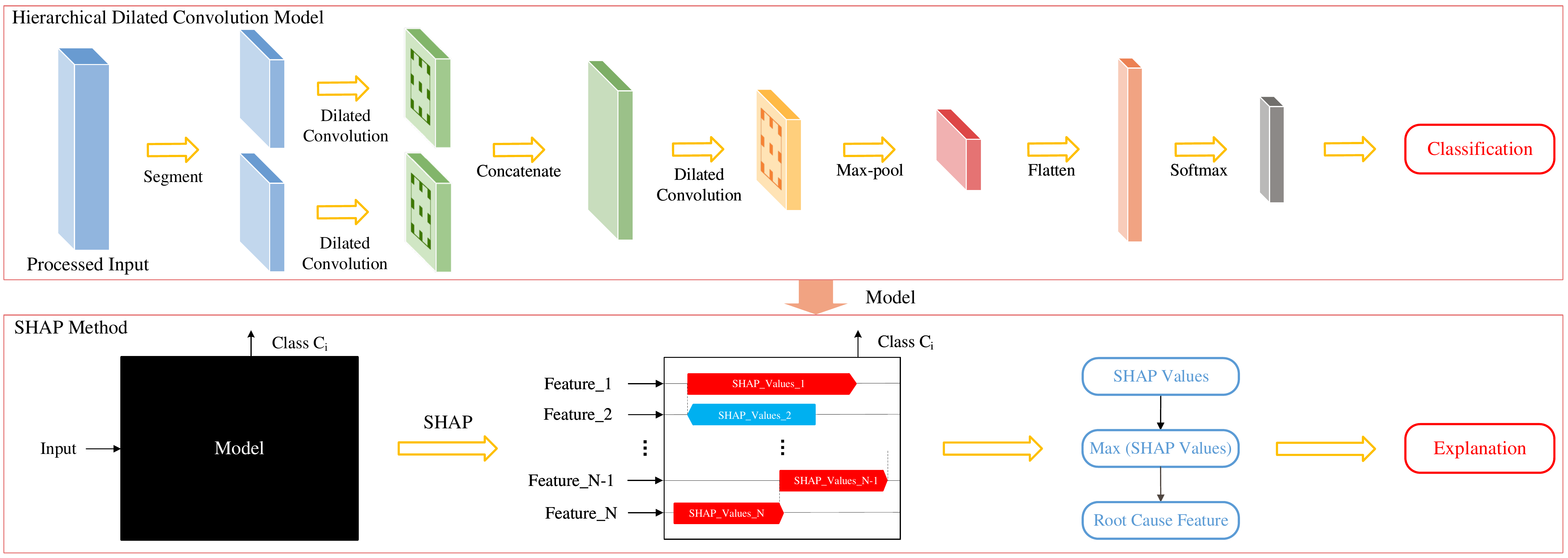}
\caption{The overall architecture of the proposed order-invariant and interpretable HDLCNN-SHAP method for chemical fault detection and diagnosis.}
\label{fig_arch}
\end{figure*}

\subsection{SHAP Method}
In this paper, we utilize the SHAP method \cite{lundberg2017unified-SHAP} to provide interpretability and obtain the root-cause features, which is based on shapley values \cite{shapley1997value-shapely} and game theory \cite{vstrumbelj2014explaining-game-theory}. It computes the shapley values for each feature of the data samples and these values indicate the contribution that the feature generates in the prediction. More specifically, a shapley value is the average marginal contribution of a feature among all possible coalitions \cite{shapley1997value-shapely}. Consider a simple linear model:
\begin{equation}
    \hat{f}(x)=\beta_{0}+\beta_{1} x_{1}+\ldots+\beta_{p} x_{p}
\end{equation}
where $x$ is a data sample with $p$ features and each $x_i$ is a feature value. $\beta_i$ is the weight of the feature $i$. The contribution $\phi_i$ of the feature $i$ on the prediction $\hat{f}(x)$ can be computed as follows:
\begin{equation}
    \phi_{i}(\hat{f})=\beta_{i} x_{i}-E\left(\beta_{i} X_{i}\right)=\beta_{i} x_{i}-\beta_{i} E\left(X_{i}\right)
\end{equation}
where $E\left(\beta_{i} X_{i}\right)$ is the mean effect estimate for the feature $i$. Then, the contribution is the difference between the feature effect and the average effect.

On this basis, SHAP defines an explanation model $g$ based on the additivity property of shapley values as follows:
\begin{equation}
\label{SHAP_equ}
    g\left(z^{\prime}\right)=\phi_{0}+\sum_{i=1}^{M} \phi_{i} z_{i}^{\prime}
\end{equation}
where $z^{\prime} \in \{0, 1\}^M$ is a one-hot vector of features, $M$ is the number of input features and $\phi_i$ is the contribution of the feature $i$. Assuming a model inputs a dataset $P$ and outputs a prediction $Y(S)$, the shapley value $\phi_i$ is computed as follows:
\begin{equation}
    \phi_{i}=\sum_{S \subseteq F \backslash\{i\}}\frac{|S| !(|F|-|S|-1) !}{|F| !}\cdot\boldsymbol{Y}
\end{equation}
\begin{equation}
    \boldsymbol{Y}=Y_{S \cup\{i\}}\left(x_{S \cup\{i\}}\right)-Y_{S}\left(x_{S}\right)
\end{equation}
where $S$ is a subset of the features and $F$ is the set of all features \cite{lundberg2017unified-SHAP}.

\section{METHODOLOGY}
\label{method}
In this paper, we propose an order-invariant and interpretable fault diagnosis method, namely HDLCNN-SHAP, based on feature clustering, dilated convolution and the SHAP method for chemical fault detection and diagnosis. The proposed method mainly contains two parts: a hierarchical dilated convolution model and an explainer based on the SHAP method. The input data is first pre-processed by the feature clustering method and then fed into the hierarchical dilated convolution model to provide classification results. Then the trained model is seen as a black box and we apply the SHAP method to interpret the model performance. The overall architecture is shown in Fig. \ref{fig_arch}. In the following subsections, we will introduce the main flow of our method in detail.

\subsection{Data Pre-processing}
\label{feature clusetering}
As shown in Fig. \ref{fig_corr}, different features may be strongly correlated which requires the ability of extracting the hidden information of the relevance among these features. In this case, a convolution layer is difficult to extract enough information within size-fixed kernels. Instead of seeking the optimal order, we cluster the correlated features before training the dilated convolution model. As a data pre-processing step, we apply hierarchical clustering to divide the features into two categories based on relevance. Compared with other clustering methods, hierarchical clustering is easy to understand and implement. More importantly, the clustering results are not affected by the input order of data. Assume that we have a set of training samples: $\boldsymbol{X}=\left\{\boldsymbol{x}_{\mathbf{1}}, \boldsymbol{x}_{\mathbf{2}}, \boldsymbol{x}_{\mathbf{3}}, \ldots, \boldsymbol{x}_{\boldsymbol{n}}\right\}$, where $\boldsymbol{x}_{\boldsymbol{i}} \in \boldsymbol{R}^{\boldsymbol{p}}$. Based on it, we build a set of the $\boldsymbol{p}$ features: $\boldsymbol{F}=\left\{\boldsymbol{f}_{\mathbf{1}}, \boldsymbol{f}_{\mathbf{2}}, \boldsymbol{f}_{\mathbf{3}}, \ldots, \boldsymbol{f}_{\boldsymbol{p}}\right\}$, where $\boldsymbol{f}_{\boldsymbol{i}} \in \boldsymbol{R}^{\boldsymbol{n}}$. Then we divide these $\boldsymbol{p}$ features into two categorizes based on the hierarchical clustering algorithm. It mainly contains the following three steps:
\begin{enumerate}
    \item Each feature $\boldsymbol{f}_{\boldsymbol{i}}$ is treated as a single cluster. Then we compute the euclidean distance $d(\boldsymbol{f}_{\boldsymbol{i}}, \boldsymbol{f}_{\boldsymbol{j}})$ between two clusters $\boldsymbol{f}_{\boldsymbol{i}}$ and $\boldsymbol{f}_{\boldsymbol{j}}$.
    \item The two closest clusters $\boldsymbol{f}_{\boldsymbol{i}}, \boldsymbol{f}_{\boldsymbol{j}}$ are merged into a single cluster $\boldsymbol{f}_{\boldsymbol{s}}$. Then $\boldsymbol{f}_{\boldsymbol{i}}, \boldsymbol{f}_{\boldsymbol{j}}$ are removed and $\boldsymbol{f}_{\boldsymbol{s}}$ is added.
    \item Iterate the previous step until there is only one cluster remaining.
\end{enumerate}

At each iteration, the distance matrix is updated to reflect the distance of the newly formed cluster $\boldsymbol{f}_{\boldsymbol{s}}$ with the remaining clusters. We use the Ward variance minimization algorithm to calculate the distance mentioned above. The distance between the newly formed cluster $\boldsymbol{f}_{\boldsymbol{s}}$ and any remaining cluster $\boldsymbol{f}_{\boldsymbol{t}}$ is defined as:
\begin{equation}
    d^{\boldsymbol{*}}(\boldsymbol{f}_{\boldsymbol{s}}, \boldsymbol{f}_{\boldsymbol{t}})=\sqrt{d_1(\boldsymbol{f}_{\boldsymbol{s}}, \boldsymbol{f}_{\boldsymbol{t}}) + d_2(\boldsymbol{f}_{\boldsymbol{s}}, \boldsymbol{f}_{\boldsymbol{t}}) - d_3(\boldsymbol{f}_{\boldsymbol{s}}, \boldsymbol{f}_{\boldsymbol{t}})}
\end{equation}
\begin{equation}
    d_1(\boldsymbol{f}_{\boldsymbol{s}}, \boldsymbol{f}_{\boldsymbol{t}}) = \frac{|\boldsymbol{f}_{\boldsymbol{t}}|+|\boldsymbol{f}_{\boldsymbol{i}}|}{T} d(\boldsymbol{f}_{\boldsymbol{t}}, \boldsymbol{f}_{\boldsymbol{i}})^{2}
\end{equation}
\begin{equation}
    d_2(\boldsymbol{f}_{\boldsymbol{s}}, \boldsymbol{f}_{\boldsymbol{t}}) = \frac{|\boldsymbol{f}_{\boldsymbol{t}}|+|\boldsymbol{f}_{\boldsymbol{j}}|}{T} d(\boldsymbol{f}_{\boldsymbol{t}}, \boldsymbol{f}_{\boldsymbol{j}})^{2}
\end{equation}
\begin{equation}
    d_3(\boldsymbol{f}_{\boldsymbol{s}}, \boldsymbol{f}_{\boldsymbol{t}}) = \frac{|\boldsymbol{f}_{\boldsymbol{t}}|}{T} d(\boldsymbol{f}_{\boldsymbol{i}}, \boldsymbol{f}_{\boldsymbol{j}})^{2}
\end{equation}
where $T=|\boldsymbol{f}_{\boldsymbol{t}}|+|\boldsymbol{f}_{\boldsymbol{i}}|+|\boldsymbol{f}_{\boldsymbol{j}}|$. Finally, we obtain two sets of features $\boldsymbol{F}_{\mathbf{1}}, \boldsymbol{F}_{\mathbf{2}}$ and each of them contains features with high correlation coefficients. The original data is reordered based on $\boldsymbol{F}_{\mathbf{1}}$ and $\boldsymbol{F}_{\mathbf{2}}$. For each sample $\boldsymbol{x}_{\boldsymbol{i}}$, the processed sample $\boldsymbol{x}_{\boldsymbol{i}}^{\boldsymbol{\prime}}$ contains $\boldsymbol{p}$ features: $\boldsymbol{F}^{\boldsymbol{\prime}}=\left\{\boldsymbol{f}_{\mathbf{1}}^{\boldsymbol{\prime}}, \ldots, \boldsymbol{f}_{\mathbf{m}}^{\boldsymbol{\prime}}, \boldsymbol{f}_{\mathbf{m+1}}^{\boldsymbol{\prime}}, \ldots, \boldsymbol{f}_{\boldsymbol{p}}^{\boldsymbol{\prime}}\right\}$, where $\left\{\boldsymbol{f}_{\mathbf{1}}^{\boldsymbol{\prime}}, \ldots, \boldsymbol{f}_{\boldsymbol{m}}^{\boldsymbol{\prime}}\right\}$ belongs to $\boldsymbol{F}_{\mathbf{1}}$ and $\left\{\boldsymbol{f}_{\mathbf{m+1}}^{\boldsymbol{\prime}}, \ldots, \boldsymbol{f}_{\boldsymbol{p}}^{\boldsymbol{\prime}}\right\}$ belongs to $\boldsymbol{F}_{\mathbf{2}}$.

\begin{figure*}[]
\centering
\includegraphics[width = \textwidth]{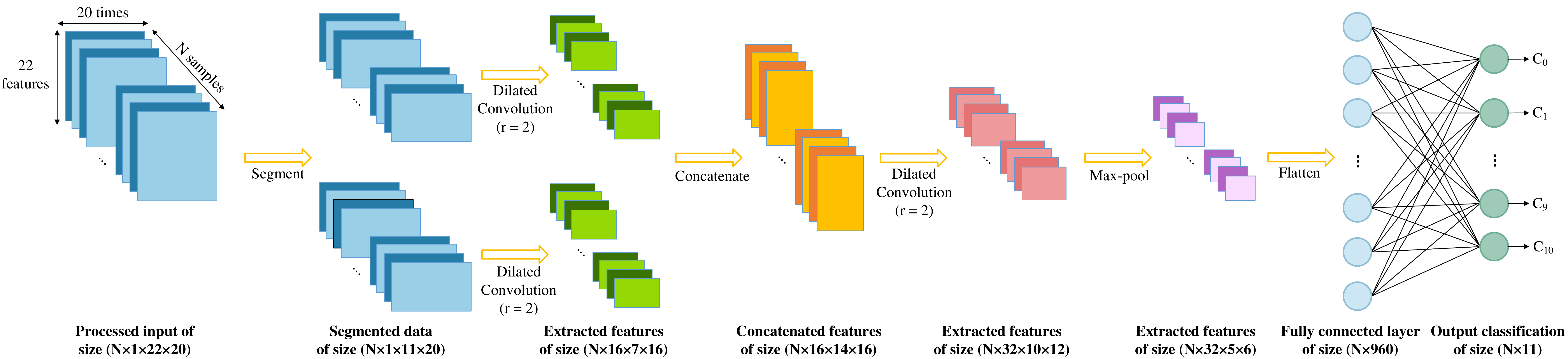}
\caption{Details of the proposed order-invariant hierarchical model based on dilated convolution.} 
\label{fig_model}
\end{figure*}

\subsection{Hierarchical Dilated Convolution Model} 
\label{HDLCNN}
After feature clustering, the processed data is then fed into the dilated convolution layers for feature extraction. To explain the model structure clearly, we take the data from TE dataset as input and define the processed input data as a set $\boldsymbol{D}=\left\{\boldsymbol{d}_{\mathbf{1}}, \boldsymbol{d}_{\mathbf{2}}, \boldsymbol{d}_{\mathbf{3}}, \ldots, \boldsymbol{d}_{\boldsymbol{n}}\right\}$, where $\boldsymbol{d}_{\boldsymbol{i}} \in \boldsymbol{R}^{\boldsymbol{22\times 20}}$. Each data sample $\boldsymbol{d}_{\boldsymbol{i}}$ has 22 features and 20 time duration. The details are shown in Fig. \ref{fig_model} and the procedures of the model are summarized as follows.
\begin{enumerate}
    \item The size of the processed data is $(N\times1\times22\times20)$ and we divide the 22 features into two segments. Since the processed data is reordered in the previous step, each segment contains highly correlated features belonging to the same cluster. Each segment has a size of $(N\times1\times11\times20)$.
    \item The segmented data is processed by a dilated convolution layer with dilation rate $r=2$. Then the hidden information about feature correlation is extracted locally and the size of the extracted features is $(N\times16\times7\times16)$.
    \item The extracted features are concatenated to obtain the entire information about two feature clusters. The size of the concatenated features is $(N\times16\times14\times16)$.
    \item The concatenated features are then processed by a dilated convolution layer with dilation rate $r=2$. This step can further extract the global information and the size of the new extracted features is $(N\times32\times10\times12)$.
    \item A max pooling layer is applied to help over-fitting and reduce the computational cost. Now the size of the extracted features is $(N\times32\times5\times6)$.
    \item The extracted features are flattened to couple information that exists vertically and horizontally. The output data of the fully connected layer has a size of $(N\times960)$.
    \item A linear layer is used to change the dimensionality of the data. Then a softmax activation function is applied to impart non-linearity into the model and output the probability distributions of the possible classes (11 in this case). The size of the final output is $(N\times11)$.
\end{enumerate}

Finally, we obtain the classification results and a trained model that requires explanation. The performance of this model is described in Section \ref{experiment} and the interpretability method is introduced in the following subsection.

\subsection{Deep SHAP Explainer}
To interpret the order-invariant hierarchical model mentioned in Section \ref{HDLCNN}, we apply an explainer based on the SHAP method which combines deep learning important features (DeepLIFT) \cite{shrikumar2017learning-deeplift} and shapley values to leverage extra knowledge about the properties of deep neural networks to improve computational performance \cite{lundberg2017unified-SHAP}.

DeepLIFT is an algorithm to compute the feature importance of the input with a given output based on back-propagation \cite{shrikumar2017learning-deeplift}. This method uses a summation-to-delta property to compute the contribution scores $C_{\Delta x_{i} \Delta y}$ for each input $x_{i}$:
\begin{equation}
    \sum_{i=1}^{n} C_{\Delta x_{i} \Delta y}=\Delta y
\end{equation}
where $y$ is the model output, $\Delta y=y-y^{0}$, $\Delta x_{i}=x_{i}-x^{0}$, $x^{0}$ refers to the reference input, and $y^{0}$ represents the reference output. Compared with Equation \ref{SHAP_equ}, if we define $\phi_{i}=C_{\Delta x_{i} \Delta y}$ and $\phi_{0}=y^{0}$, then DeepLIFT approximates SHAP values for linear models.

Deep SHAP takes DeepLIFT as a compositional approximation of SHAP values and recursively passes the multipliers of DeepLIFT backwards through the network \cite{lundberg2017unified-SHAP}. Deep SHAP explainer can effectively achieve linearization by combining the SHAP values computed for smaller components into SHAP values for the whole model. Therefore, we can quantify the contribution of each feature from each data sample to obtain local explanation. Based on the feature contribution of each data sample, we further interpret the model globally by calculating the average value of data samples for each feature, which can be mathematically described as follows:
\begin{equation}
    \Phi_{i} = \frac{1}{n}\sum_{j=1}^{n} \phi_{i}(x_j)
\end{equation}
where $\phi_{i}(x_j)$ refers to the contribution of the feature $i$ of the input $x_j$. Finally, we identify the root-cause feature $\boldsymbol{\gamma}$ as the one with the highest contribution.
\begin{equation}
    \boldsymbol{\gamma}=\arg\max(\Phi_{i})
\end{equation}

\subsection{The Entire Fault Detection and Diagnosis Procedure}
Based on the description above, the entire order-invariant and interpretable fault detection and diagnosis procedure via feature clustering, hierarchical dilated convolution model and SHAP method mainly consists of six steps:
\begin{enumerate}
    \item Collecting the monitored variables via sensors in chemical processes. Obtaining the training set from the collected samples and normalizing the data.
    \item Processing the training samples based on hierarchical clustering method. The features are reordered according to the clustering results and the highly correlated features are closer in the processed data.
    \item Training the hierarchical dilated convolution model with the processed data. Storing the model parameters for later usage.
    \item Acquiring online samples and processing them in the same way as mentioned above.
    \item Restoring the trained model and classifying the new sample into a normal or fault type.
    \item If a fault happens, identifying the corresponding root-cause feature based on SHAP method. Feature contribution is computed and the one with highest contribution is considered as the root-cause feature.
\end{enumerate}

\begin{table*}[]
\centering
\caption{Binary Fault detection accuracy of the selected 10 faults}
\label{binary-accuracy}
\setlength{\tabcolsep}{1.49mm}{
\begin{tabular}{|c|cc|cc|cc|cc|cc|c|c|c|c|c|c|}
\hline
\multirow{2}{*}{Fault ID} & \multicolumn{2}{c|}{PCA}                      & \multicolumn{2}{c|}{KPCA}            & \multicolumn{2}{c|}{KDPCA}                     & \multicolumn{2}{c|}{KDICA}  & \multicolumn{2}{c|}{MLPP}          & \multirow{2}{*}{DSAE} & \multirow{2}{*}{VS-SVDD} & \multirow{2}{*}{MBTCN} &  \multirow{2}{*}{CNN} & \multirow{2}{*}{DLCNN} & \multirow{2}{*}{HDLCNN} \\ \cline{2-11}
                          & \multicolumn{1}{c|}{SPE}  & T$^{2}$ &  \multicolumn{1}{c|}{SPE}   & T$^{2}$
                          & \multicolumn{1}{c|}{SPE}   & T$^{2}$ &  \multicolumn{1}{c|}{SPE}   & T$^{2}$ & \multicolumn{1}{c|}{SPE}   & T$^{2}$ &                          &            &       &           &     &       \\ \hline
1                   & \multicolumn{1}{c|}{99.5} & 99.1     & \multicolumn{1}{c|}{100.0} & 99.3    & \multicolumn{1}{c|}{100.0} & 99.5        & \multicolumn{1}{c|}{100.0} & 100.0 & \multicolumn{1}{c|}{99.7} & 100.0   & 99.3 & 99.0  & 100.0              & 99.4        & 98.8        & 100.0                 \\ \hline
2                   & \multicolumn{1}{c|}{98.4} & 98.5     & \multicolumn{1}{c|}{99.0}  & 95.3    & \multicolumn{1}{c|}{99.1}  & 98.3            & \multicolumn{1}{c|}{98.5}  & 98.8  & \multicolumn{1}{c|}{98.9}  & 99.8  & 96.8 & 98.0     & 99.0      & 99.1         & 99.1           & 98.7                  \\ \hline
3                   & \multicolumn{1}{c|}{0.6}  & 3.6      & \multicolumn{1}{c|}{6.8}   & 9.0     & \multicolumn{1}{c|}{9.6}   & 4.4             & \multicolumn{1}{c|}{19.4}  & 19.8  & \multicolumn{1}{c|}{23.8}  & 39.6  & 67.4  & 42.0    & 85.4           & 85.7        & 92.4        & 97.6                  \\ \hline
8                   & \multicolumn{1}{c|}{96.8} & 97.4     & \multicolumn{1}{c|}{97.9}  & 97.4    & \multicolumn{1}{c|}{97.8}  & 97.6           & \multicolumn{1}{c|}{97.8}  & 99.4  & \multicolumn{1}{c|}{100.0}  & 98.7  & 87.0  & 98.0     & 89.0         & 96.7       & 97.7           & 95.9                  \\ \hline
10                  & \multicolumn{1}{c|}{15.4} & 36.7     & \multicolumn{1}{c|}{52.5}  & 48.6    & \multicolumn{1}{c|}{63.5}  & 42.6            & \multicolumn{1}{c|}{80.6}  & 92.9  & \multicolumn{1}{c|}{71.3}  & 94.2  & 68.3  & 73.0     & 86.6         & 93.7        & 93.9          & 95.6                  \\ \hline
11                  & \multicolumn{1}{c|}{63.8} & 41.4     & \multicolumn{1}{c|}{77.6}  & 51.0    & \multicolumn{1}{c|}{91.0}  & 33.6            & \multicolumn{1}{c|}{81.4}  & 90.3  & \multicolumn{1}{c|}{93.6}  & 95.6  & 81.3   & 98.0     & 99.5        & 93.4         & 93.0        & 94.5                  \\ \hline
12                  & \multicolumn{1}{c|}{92.5} & 98.5     & \multicolumn{1}{c|}{98.5}  & 98.9    & \multicolumn{1}{c|}{99.1}  & 99.1           & \multicolumn{1}{c|}{99.7}  & 100.0  & \multicolumn{1}{c|}{99.6}  & 100.0 & 94.1   & 100.0    & 96.5        & 81.7       & 82.7         & 90.7                  \\ \hline
13                  & \multicolumn{1}{c|}{95.0} & 94.3     & \multicolumn{1}{c|}{95.2}  & 94.3    & \multicolumn{1}{c|}{95.4}  & 96.3           & \multicolumn{1}{c|}{95.9}  & 95.9  & \multicolumn{1}{c|}{96.5}  & 91.2   & 78.1   & 95.0    & 95.6       & 96.3        & 96.8         & 96.9                  \\ \hline
14                  & \multicolumn{1}{c|}{99.9} & 98.8     & \multicolumn{1}{c|}{100.0} & 99.6    & \multicolumn{1}{c|}{100.0} & 99.9         & \multicolumn{1}{c|}{100.0} & 100.0 & \multicolumn{1}{c|}{100.0} & 100.0  & 99.6   & 100.0    & 100.0     & 100.0      & 100.0            & 100.0                 \\ \hline
20                  & \multicolumn{1}{c|}{42.3} & 34.0     & \multicolumn{1}{c|}{59.8}  & 49.1    & \multicolumn{1}{c|}{66.8}  & 51.5          & \multicolumn{1}{c|}{72.7}  & 83.9  & \multicolumn{1}{c|}{86.7}  & 93.6  & 78.6  & 78.0    &    90.1    & 97.2          & 97.3          & 96.4                  \\ \hline
Average                   & \multicolumn{1}{c|}{70.4} & 70.2    & \multicolumn{1}{c|}{78.7}  & 74.2    & \multicolumn{1}{c|}{82.2}  & 72.3           & \multicolumn{1}{c|}{84.6}  & 88.1  & \multicolumn{1}{c|}{87.0}  & 91.3  & 85.1    & 88.1  & 94.2        & 94.3         & 95.2            & \textbf{96.6}         \\ \hline
\end{tabular}}
\end{table*}

\section{EXPERIMENT STUDY}
\label{experiment}
In this paper, we use the TE dataset to verify the effectiveness of the proposed method. It simulates actual chemical processes and is widely used as a benchmark in chemical fault detection and diagnosis \cite{amin2018process-te-1-15,peng2021towards-te-peng,yu2015nonlinear-te-10}. There are totally 21 types of faults and 22 continuous measured variables in this dataset. For the training set, it has 980 samples including 500 samples in the normal case and 480 samples in the case of failure for each fault type. For the test set, it has 960 samples including 160 normal samples and 800 fault samples for each fault type. 
Details are described in the following subsections.

\begin{figure}[]
\centering
\includegraphics[width = 6.5cm]{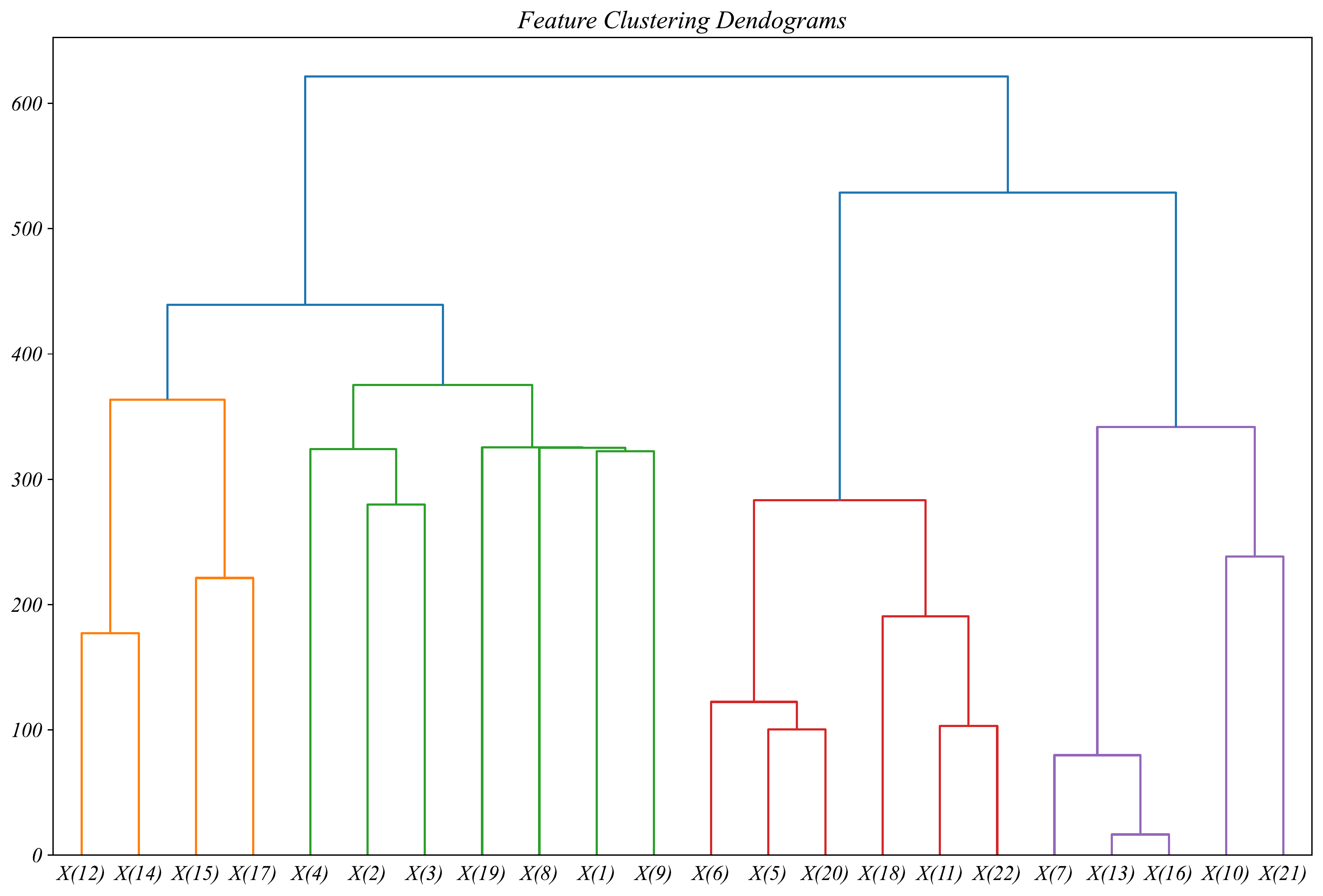}
\caption{The feature clustering dendrogram of the 22 features in TE dataset.} 
\label{fig_cluster}
\end{figure}

\subsection{Experiment Setup}
The downloaded TE dataset has a sampling period of 180 seconds, leading to few data samples for training and test. Therefore, current CNN methods use simulation model to generate more data samples for feature extraction. Similarly, we refer to the simulation method from \cite{peng2020cost} on MATLAB to obtain more data for classification. The sampling period is set to 36 seconds (100 samples/h). The simulator runs for 48 hours in the normal state, then 4800 training normal samples are collected. For each fault type, the simulator runs for 48 hours to collect 4800 training fault samples. For the testing data of each fault, the simulator runs for 8 hours in the normal state at the beginning to collect 800 test normal samples. Then a fault disturbance is introduced and the simulator continues to run for 40 hours to collect 4000 test fault samples. Next, the collected data is processed in the range [0,1] to eliminate the adverse effects caused by singular data. To extract the features in both spatial and temporal domains, each data sample is reshaped into a 2-D array with 22 features and 20 time duration.

To demonstrate the performance of our proposed model on the chemical fault detection and diagnosis, we select Fault 1, Fault 2, Fault 3, Fault 8, Fault 10, Fault 11, Fault 12, Fault 13, Fault 14, and Fault 20 for binary and multi-class fault detection and diagnosis. Fault 10 and Fault 11 are chose for root cause analysis since their root-cause features are proven and widely used. The corresponding true root-cause features are X(18) and X(9) \cite{amin2018process-te-1-15, yu2015nonlinear-te-10}.

\subsection{Feature Clustering}
As shown in Fig. \ref{fig_tabular}, highly correlated features are closer in the optimal order. Although it is hard to obtain the optimal order, we can cluster the features with higher correlation to achieve similar effect. As described in Section \ref{feature clusetering}, we divide the 22 features into two categorizes based on the correlation. The corresponding hierarchical clustering dendrogram is shown in Fig. \ref{fig_cluster}. We can see that the first category includes 11 features which are X(1), X(2), X(3), X(4), X(8), X(9), X(12), X(14), X(15), X(17) and X(19). The second category also includes 11 features which are X(5), X(6), X(7), X(10), X(11), X(13), X(16), X(18), X(20), X(21) and X(22). Refer to the correlation coefficients of the 22 features shown in Fig. \ref{fig_corr}, we can see that the highly correlated features are classified into the same cluster.

\begin{figure}[]
\centering
\includegraphics[width = 6.5cm]{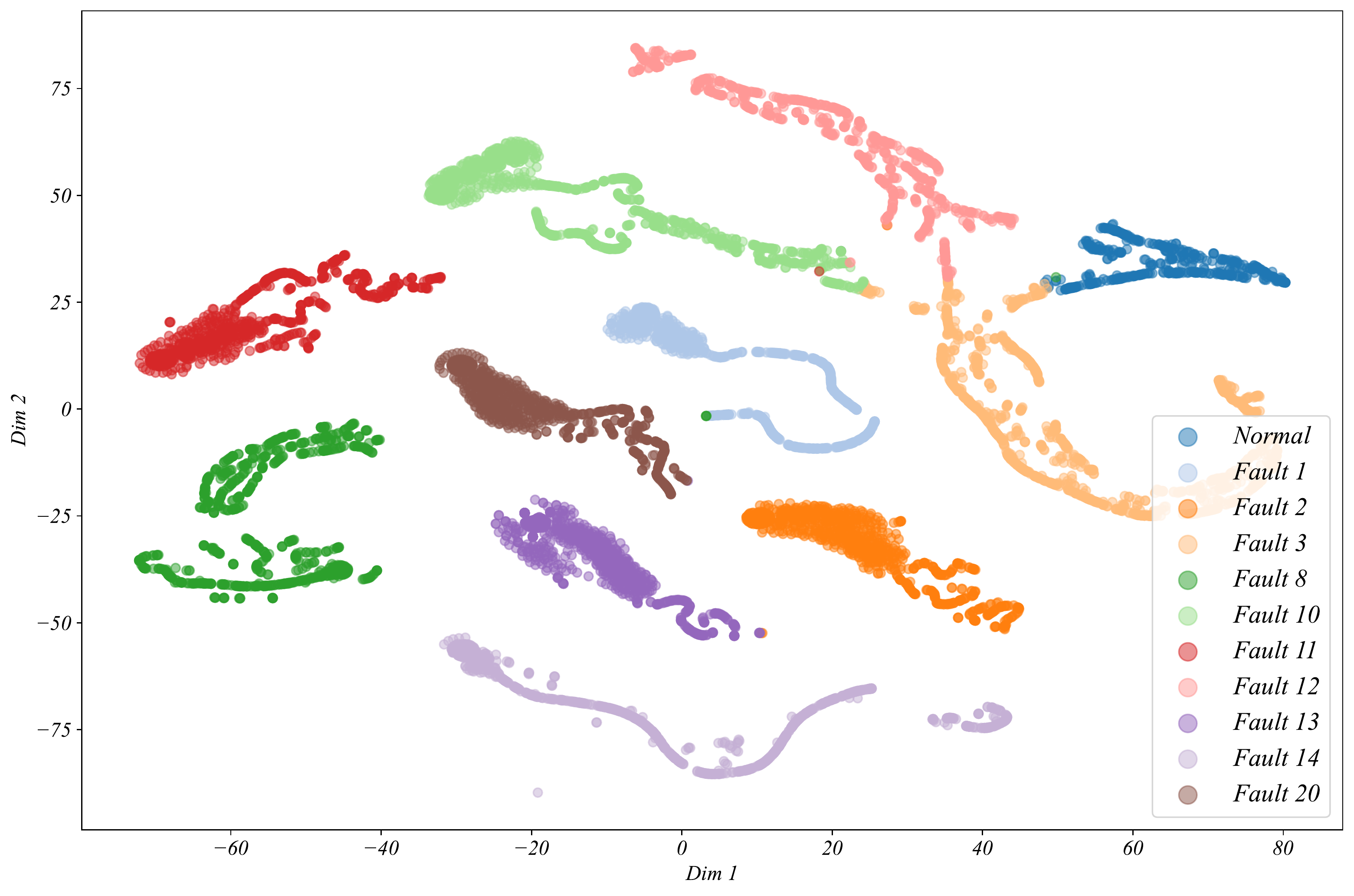}
\caption{The t-SNE embedding of the extracted features of our proposed HDLCNN model.} 
\label{fig_tsne}
\end{figure}

\subsection{Contrast and Ablation Experiments}
Firstly, we evaluate the proposed method for binary fault detection and diagnosis. In this case, only one fault is considered at a time. To show the efficiency, the proposed HDLCNN model is compared with the existing data-driven methods including principal component analysis (PCA),  kernel principal component analysis (KPCA), integrated kernel dynamic principal component analysis (KDPCA), kernel dynamic independent component analysis (KDICA) \cite{fan2014fault-comparison}, modified locality preserving projection (MLPP) \cite{shah2022modified-mlpp}, denoising sparse autoencoder (DSAE) \cite{peng2021towards-te-peng}, variable selection and support vector data description (VS-SVDD) \cite{cai2022relevant-RVS-SVDD} and multi-block temporal convolutional network (MBTCN) \cite{he2021multiblock-mbtcn}. As shown in Table. \ref{binary-accuracy}, our proposed model results in the highest average fault detection rate, which is marked bold. The following experiment results are marked in a similar way. As ablation experiments, we compare CNN, DLCNN and HDLCNN. CNN is the baseline model with traditional convolution layers and DLCNN contains dilated convolution layers without hierarchical feature clustering. HDLCNN is our proposed method involving both hierarchical feature clustering and dilated convolution layers. It is obvious that the average fault detection accuracy of DLCNN is increased by 1.0\% compared to CNN, which shows the effect of dilated convolution. In addition, hierarchical feature clustering is proved instrumental since the average fault detection accuracy of HDLCNN is increased by 1.5\% compared to DLCNN.



\begin{table}[]
\centering
\caption{Multi-class Fault detection accuracy of the selected 10 faults}
\setlength{\tabcolsep}{3.2mm}{
\begin{tabular}{|c|c|c|c|c|c|}
\hline
Fault ID & PCA  & DSAE & CNN & DLCNN  & HDLCNN          \\ \hline
0        & 5.1  & 51.3 & 85.3 & 92.1 & 99.6          \\ \hline
1        & 51.8 & 98.4 & 99.2 & 99.7 & 97.7         \\ \hline
2        & 78.5 & 97.8 & 94.6 & 95.0 & 97.2          \\ \hline
3        & 25.9 & 14.1 & 94.4 & 100.0 & 100.0         \\ \hline
8        & 4.1  & 44.8 & 89.6 & 93.0 & 94.5          \\ \hline
10       & 8.6  & 36.4 & 95.3 & 99.4 & 98.8         \\ \hline
11       & 16.4 & 45.5 & 90.3 & 93.1 & 93.0          \\ \hline
12       & 12.2 & 68.4 & 73.4 & 95.5 & 93.9          \\ \hline
13       & 45.4 & 23.8 & 92.2 & 75.2 & 93.3          \\ \hline
14       & 50.8 & 97.9 & 100.0 & 100.0 & 100.0         \\ \hline
20       & 41.5 & 72.4 & 89.9 & 91.3 & 92.6          \\ \hline
Average  & 30.9 & 59.2 & 91.3 & 94.0 & \textbf{96.4} \\ \hline
\end{tabular}}
\label{table_11}
\end{table}

\begin{table}[]
\centering
\caption{Accuracy of CNN, DLCNN and HDLCNN}
\label{acc-te}
\setlength{\tabcolsep}{4mm}{
\begin{tabular}{|c|c|c|c|}
\hline
  & CNN & DLCNN      & HDLCNN \\ \hline
Close-correlated Order & 91.3   & 94.0       &   96.4    \\ \hline
Separate-correlated Order & 85.8   & 91.8       &   96.0   \\ \hline
Difference   & 5.5    & 2.2        &   0.4    \\ \hline
\end{tabular}}
\end{table}

Then, to explore the performance of the proposed method for multi-class fault detection and diagnosis, we combine the normal case with the selected 10 types of faults. As shown in Fig. \ref{fig_tsne}, the extracted features of HDLCNN are visualized. Specifically, we utilize t-SNE method to reduce the dimension of the features to 2, and then plot them by class. It is obvious that the embedding of the extracted features belonging to different classes are separate. Therefore, it is unsurprising that the softmax layer can get accurate classification results. PCA, DSAE, CNN and DLCNN are selected as comparison methods. As shown in Table. \ref{table_11}, the proposed model achieves the highest average fault detection rate. Similarly, we consider CNN, DLCNN and HDLCNN as ablation experiments. Due to the dilated convolution, the average fault detection accuracy of DLCNN is increased by 3.0\% compared to CNN. And with hierarchical feature clustering, the average fault detection accuracy of HDLCNN is increased by 2.6\% compared to DLCNN.


\subsection{Ablation Experiments of Sensitivity to Feature Order}
\begin{figure}[]
\centering
\setcounter{subfigure}{0}
\subfigure[Close-correlated Order]
{
 \centering
 \includegraphics[width = 4cm]{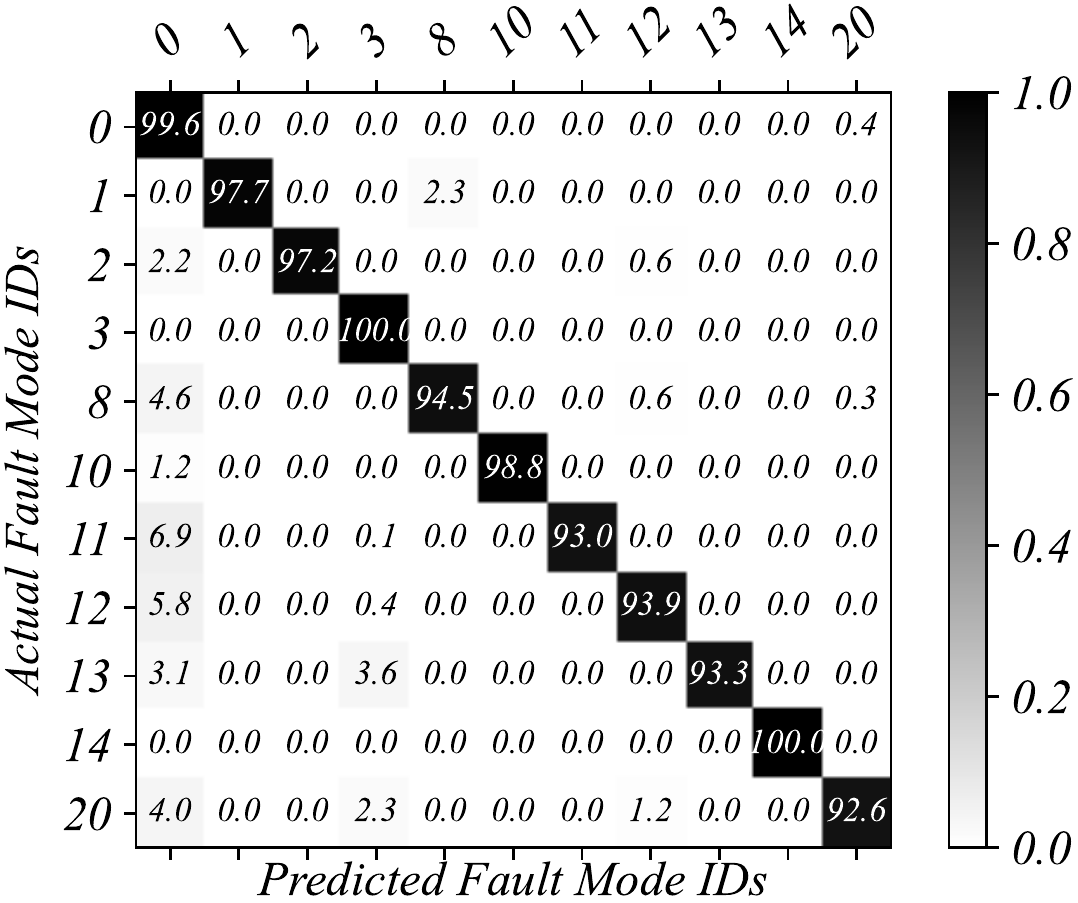}
}
\subfigure[Separate-correlated Order]
{
 \centering
 \includegraphics[width = 4cm]{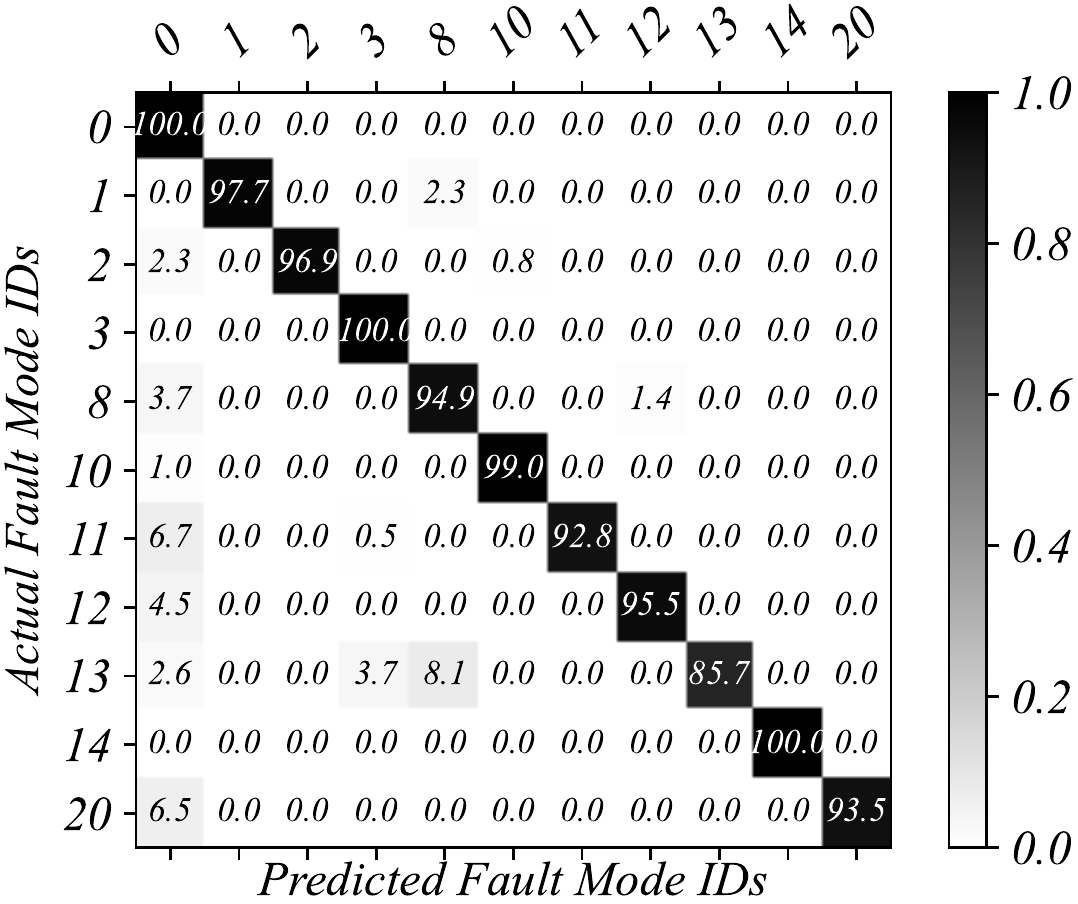}
}
\caption{The confusion matrices of HDLCNN.} 
\label{fig_confusion_matrix}
\end{figure}

To demonstrate the ability of our proposed HDLCNN model to process tabular data with features of arbitrary order, we find the separate-correlated order of features and compare to the close-correlated order. More specifically, we consider the separate-correlated order as [X(21), X(8), X(4), X(3), X(15), X(16), X(2), X(6), X(20), X(13), X(17), X(18), X(9), X(1), X(7), X(10), X(5), X(14), X(19), X(11), X(12), X(22)], in which highly correlated features are separated. And the close-correlated order is formed in an opposite way. In this case, the average multi-class fault detection accuracy of CNN, DLCNN and HDLCNN is shown in Table. \ref{acc-te}. We see that the performance of CNN and DLCNN is influenced by the order of the features and the differences are 5.5\% and 2.2\% respectively. This proves that dilated convolution can extract more information and weaken the effect of the feature order. Further, HDLCNN is order-invariant and the difference is only 0.4\%, which confirms the effect of hierarchical feature clustering. In a word, the ablation experiments demonstrate that our proposed method can effectively process tabular data with features of arbitrary order without seeking the optimal order. The confusion matrices obtained by HDLCNN are illustrated in Fig. \ref{fig_confusion_matrix}.

\begin{figure}[]
\centering
\setcounter{subfigure}{0}
\subfigure[Fault 10]
{
 \centering
 \includegraphics[width=8cm]{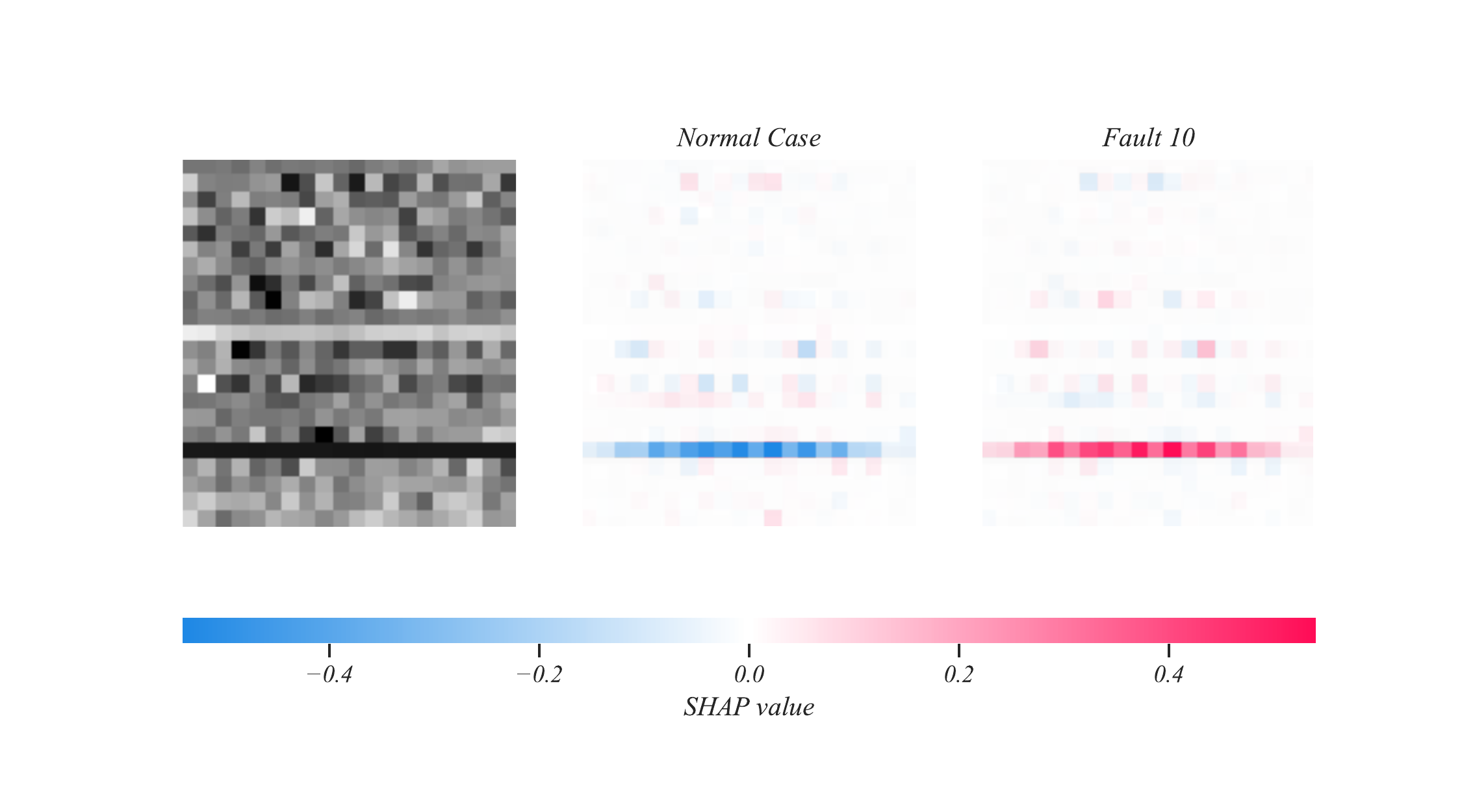}
}

\subfigure[Fault 11]
{
 \centering
 \includegraphics[width=8cm]{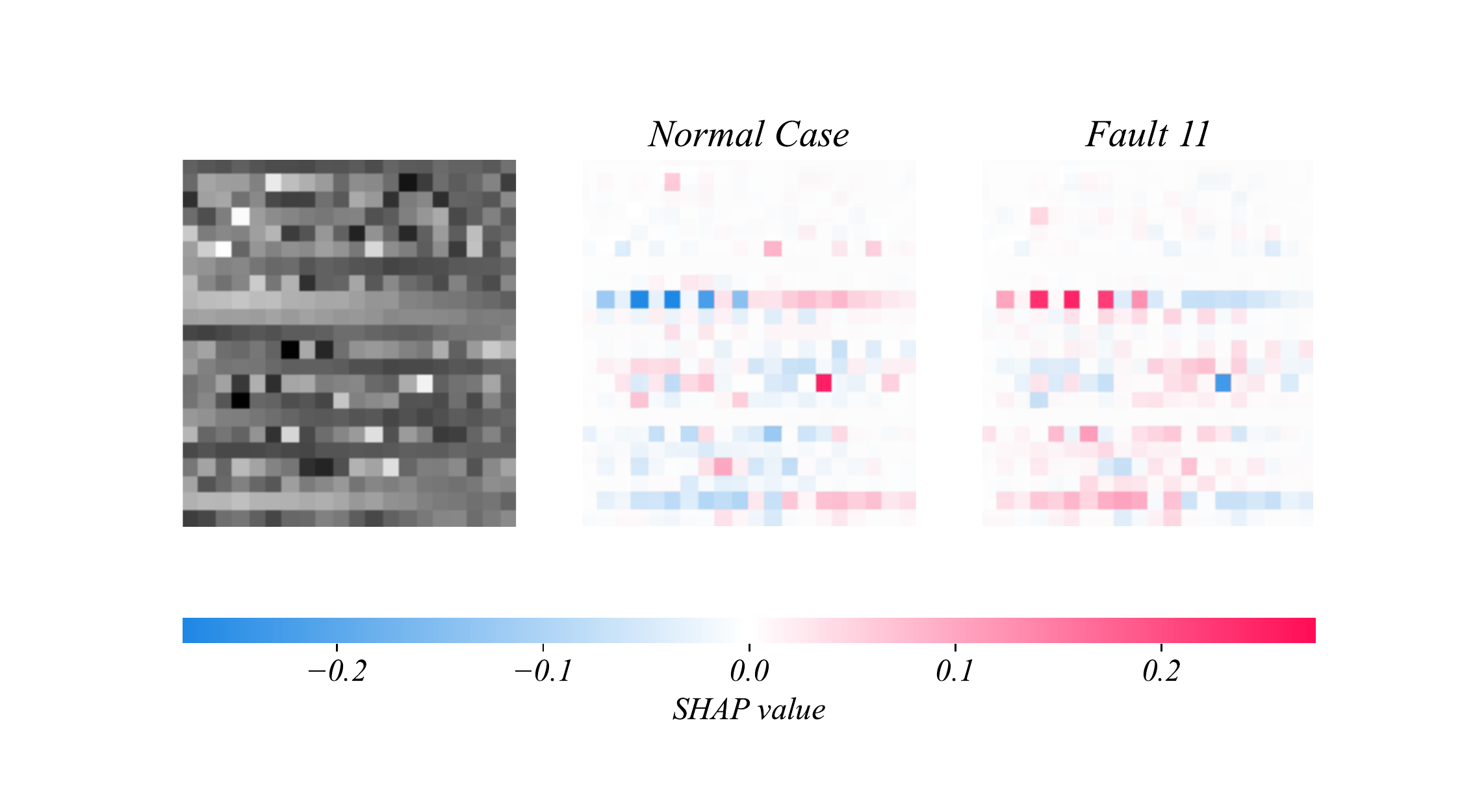}
}
\caption{The visualization of SHAP values of a single sample of Fault 10 and 11.}
\label{shap_fig_root_cause}
\end{figure}

\begin{figure}[]
\centering
\setcounter{subfigure}{0}
\subfigure[Fault 10]
{
 \centering
 \includegraphics[width=4cm]{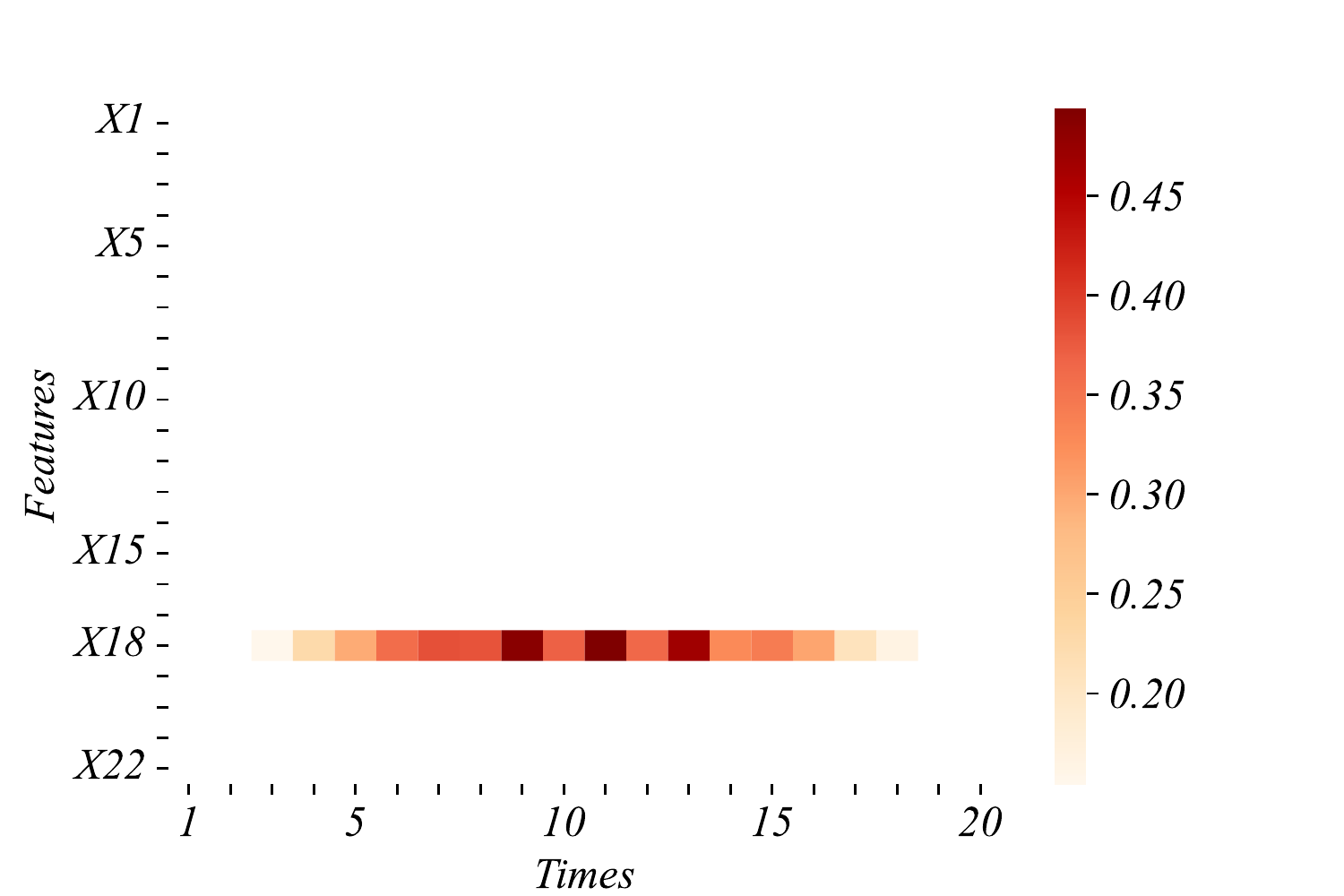}
}
\subfigure[Fault 11]
{
 \centering
 \includegraphics[width=4cm]{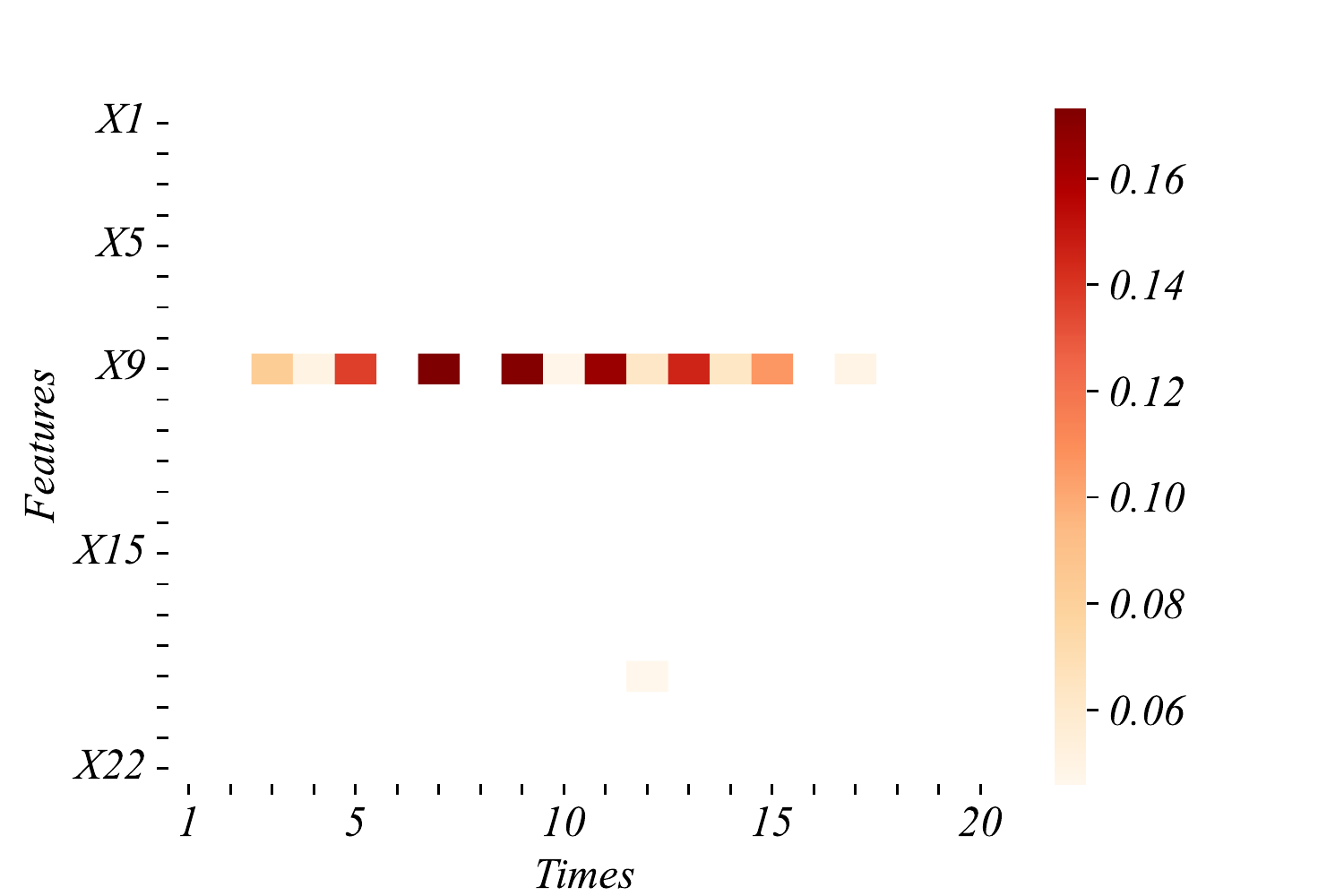}
}
\caption{The heatmap of a single sample of Fault 10 and 11.}
\label{heatmap_fig_root_cause}
\end{figure}

\subsection{Local and Global Explanation}
We further analyze the feature contribution and obtain the root-cause features based on SHAP values. Fault 10 and Fault 11 are selected for root cause analysis, and the corresponding true root-cause features are X(18) and X(9), respectively. For Fault 10, the stripper temperature (X(18)) is directly affected because of the random variation of temperature in feed C \cite{yu2015nonlinear-te-10}. And for Fault 11, the random variation in reactor cooling water inlet temperature results in abnormal behaviour of the reactor temperature (X(9)) \cite{amin2018process-te-1-15}.

\begin{figure}[]
\centering
\setcounter{subfigure}{0}
\subfigure[Fault 10]
{
 \centering
 \includegraphics[width=4cm]{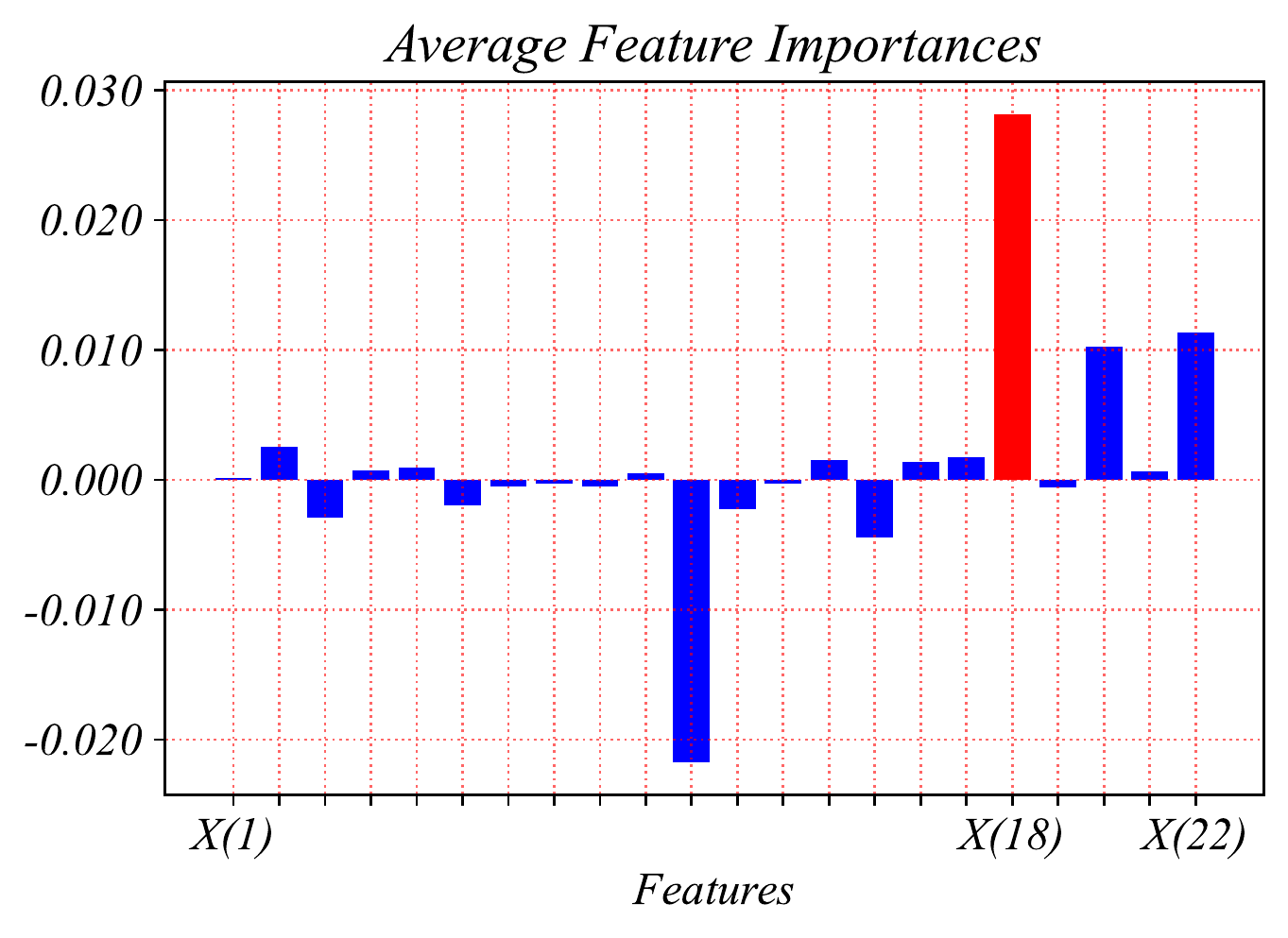}
}
\subfigure[Fault 11]
{
 \centering
 \includegraphics[width=4cm]{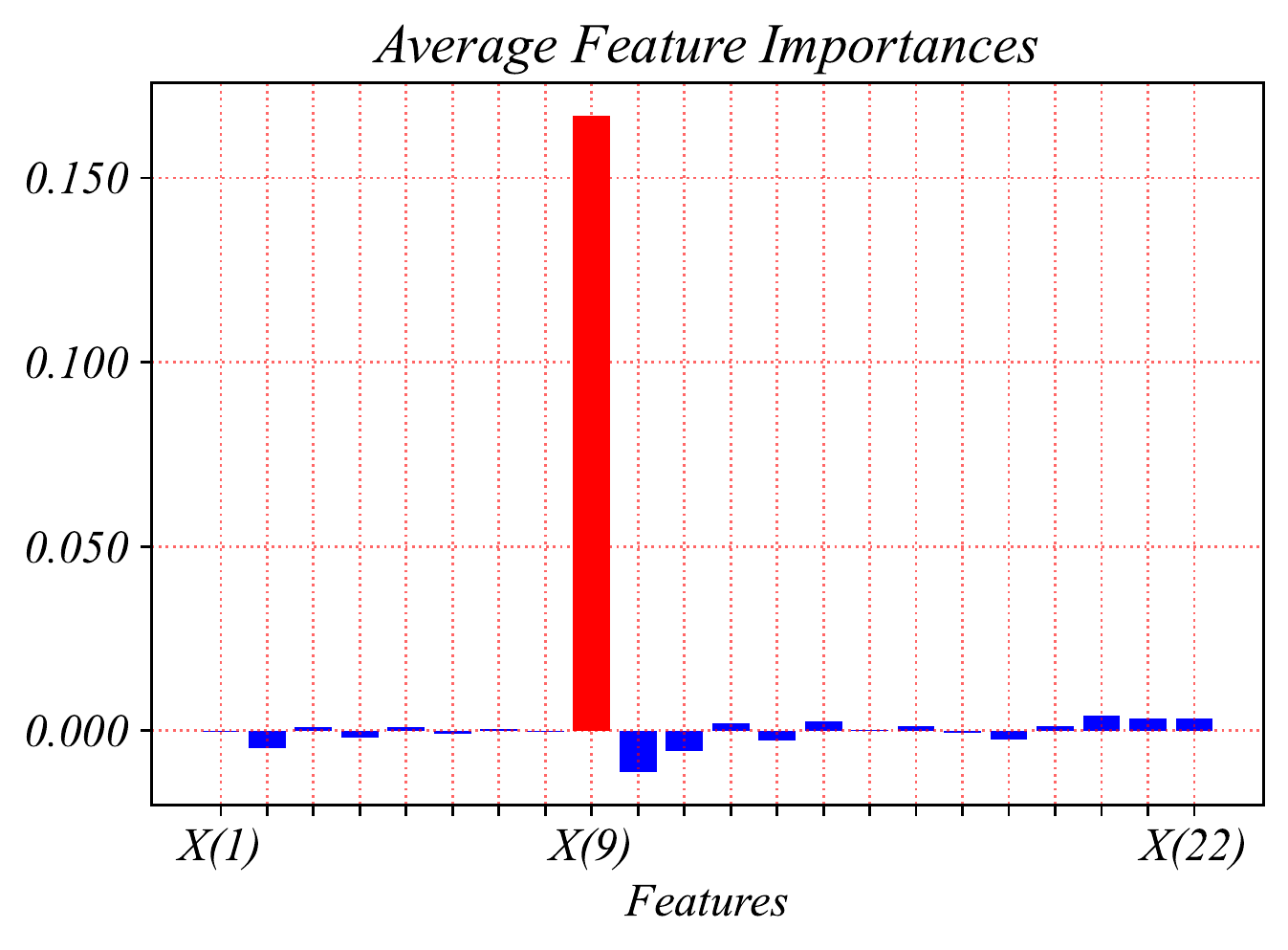}
}
\caption{The average feature importance of Fault 10 and 11.}
\label{fig_root_cause}
\end{figure}

First, local explanation is provided to indicate the feature contribution to the prediction from a single sample. Fig. \ref{shap_fig_root_cause} shows the visualization of the SHAP values of a single sample. The left column indicates the gray image of the sample, the middle column shows the SHAP values of classifying this sample to the normal case, and the right column shows the SHAP values of classifying this sample to the case of failure. Red pixels indicate high SHAP values and blue pixels denote low SHAP values. High SHAP value means great feature contribution of this sample to be classified to the corresponding type of fault. The corresponding heatmaps are shown in Fig. \ref{heatmap_fig_root_cause}. It is obviously that the most important features are X(18) and X(9) for Fault 10 and Fault 11 respectively, which are also the true root-cause features.

\begin{figure}[]
\centering
\setcounter{subfigure}{0}
\subfigure[Fault 10]
{
 \centering
 \includegraphics[width = 4cm]{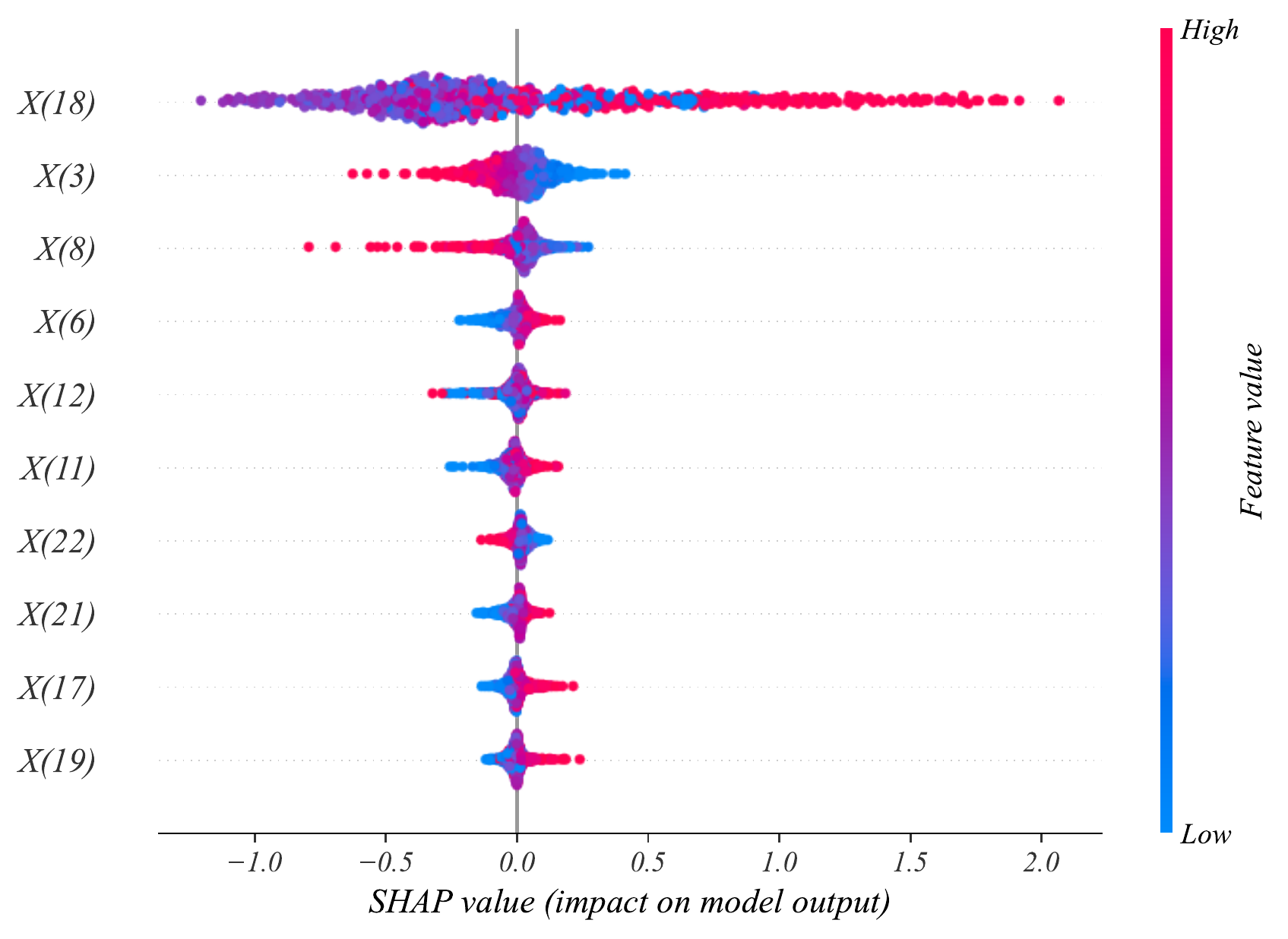}
}
\subfigure[Fault 11]
{
 \centering
 \includegraphics[width = 4cm]{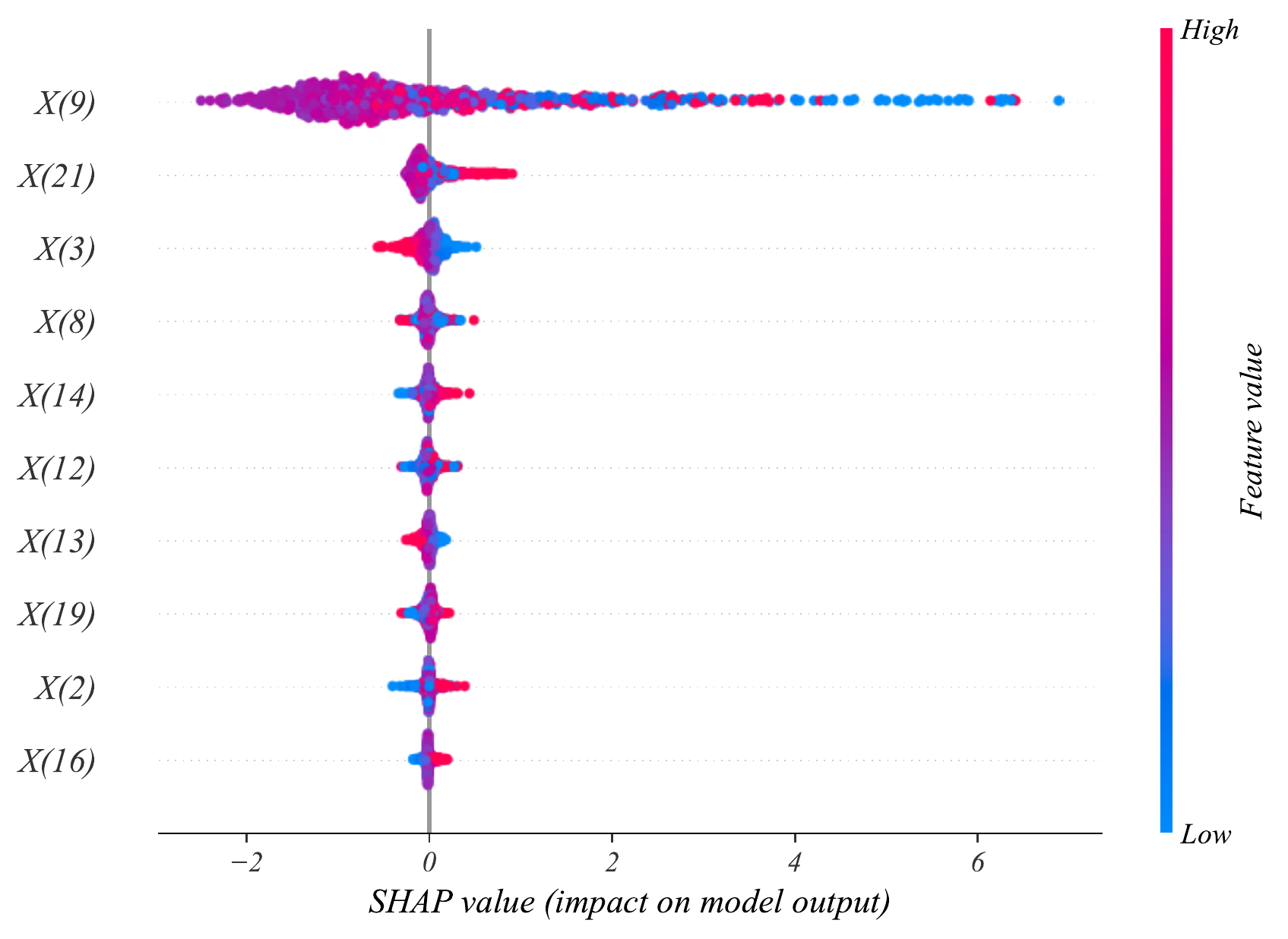}
}
\caption{The SHAP values of the top 10 features of Fault 10 and 11.} 
\label{summary_fig_root_cause}
\end{figure}

Then, global explanation is described to show the feature contribution of the overall dataset. We compute the average feature contribution for each sample and consider the feature with highest importance as the corresponding root-cause feature. Fig. \ref{fig_root_cause} shows the average feature importance and the features with highest importance are marked red. We see that the most important features are X(18) and X(9) for Fault 10 and Fault 11 respectively, which are also the true root-cause features. Fig. \ref{summary_fig_root_cause} denotes the relationship between the measured feature values and the corresponding SHAP values. For Fault 10, high measured values of X(18) obviously correspond high SHAP values which means high stripper temperature (X(18)) may be the main cause of the failure. On the contrary, for Fault 11, we see that low measured values of X(9) mainly refer to high SHAP values which reminds us to pay attention to the reduction of reactor temperature (X(9)).

\section{Conclusion}
\label{conclusion}
In this paper, we propose an order-invariant and interpretable HDLCNN-SHAP method for chemical fault detection and diagnosis based on feature clustering, dilated convolution and SHAP method. The ability to detect faults and obtain the root-cause features is essential for fault detection and diagnosis methods in real chemical processes. Comparing with the existing methods, our proposed method can effectively process tabular data with features of arbitrary order without seeking the optimal order. In addition, root-cause features are precisely identified without any human supervision. The proposed method is evaluated on a simulation dataset based on an actual chemical process. Experimental results show that the proposed method achieves better performance for both binary and multi-class fault detection and diagnosis compared with other popular data-driven methods. Moreover, the proposed method is order-invariant, which results in insensitivity to the order of the features. Local and global explanation are further described to obtain the root-cause features. In our future work, we will focus on more practical and complex fault detection problems. Simultaneous-fault diagnosis is a common problem in real applications and the problem of faults with multiple root-cause features is consequential as well. Besides, incomplete and high-dimensional datasets are worth of investigation for solving fault detection problems in real-world chemical processes.

\ifCLASSOPTIONcaptionsoff
  \newpage
\fi



%

\bibliographystyle{IEEEtran}%
\bibliography{ref.bib}
\vspace{12pt}
%




\end{document}